\documentclass{article}
\usepackage{ijcai22}
\usepackage{times}

\usepackage{soul}
\usepackage{url}
\usepackage[hidelinks]{hyperref}
\usepackage[utf8]{inputenc}
\usepackage[small]{caption}
\usepackage{graphicx}
\usepackage{booktabs}
\usepackage[table,xcdraw]{xcolor}
\usepackage[margin=1in]{geometry}
\usepackage[utf8]{inputenc}
\usepackage[ruled,vlined,linesnumbered]{algorithm2e}
\usepackage{amsthm}
\usepackage{amsmath}
\usepackage{amssymb}
\usepackage{appendix}
\usepackage{authblk}
\usepackage{bbm}
\usepackage{csquotes}
\usepackage{caption}
\usepackage{enumitem}
\usepackage{float}
\usepackage{subcaption}
\MakeOuterQuote{"}
\usepackage{natbib}
\usepackage{mathtools}
\usepackage{tikz}
\usetikzlibrary{positioning, fit, calc, shapes}

\newcommand{\C}{\mathcal{C}}
\newcommand{\R}{\mathbb{R}}

\newcommand{\bigO}{\mathcal{O}}
\newcommand{\mP}{\mathcal{P}}
\DeclareMathOperator{\E}{\mathbb{E}}

\newcommand{\U}{\mathcal{U}}

\newcommand{\rop}[1]{\mathcal{R}\left\{#1\right\}}
\newcommand{\rrop}[1]{\mathcal{R}^2\left\{#1\right\}}
\newcommand{\ropt}[2]{\mathcal{R}_{#1}\left\{#2\right\}}
\newcommand{\rropt}[2]{\mathcal{R}_{#1}^2\left\{#2\right\}}

\newtheorem{theorem}{Theorem}[section]

\newtheorem{lemma}{Lemma}[section]

\SetKwInOut{Input}{Input}
\SetKwInOut{Output}{Output}

\usepackage{etoolbox}

\makeatletter

\patchcmd{\NAT@test}{\else \NAT@nm}{\else \NAT@nmfmt{\NAT@nm}}{}{}

\DeclareRobustCommand\citepos
  {\begingroup
   \let\NAT@nmfmt\NAT@posfmt
   \NAT@swafalse\let\NAT@ctype\z@\NAT@partrue
   \@ifstar{\NAT@fulltrue\NAT@citetp}{\NAT@fullfalse\NAT@citetp}}

\let\NAT@orig@nmfmt\NAT@nmfmt
\def\NAT@posfmt#1{\NAT@orig@nmfmt{#1's}}

\title{Non-Convex Optimization with Spectral Radius Regularization}

\author[1]{Adam Sandler}
\affil[1]{Engineering Sciences and Applied Mathematics, Northwestern University}

\author[2]{Diego Klabjan}
\affil[2]{Industrial Engineering and Management Sciences, Northwestern University}
            
\author[3]{Yuan Luo}
\affil[3]{Preventive Medicine (Health and Biomedical Informatics), Northwestern University}

\begin{document}

\maketitle

\begin{abstract}
We develop regularization methods to find flat minima while training deep neural networks. These minima generalize better than sharp minima, yielding models outperforming baselines on real-world test data (which may be distributed differently than the training data). Specifically, we propose a method of regularized optimization to reduce the spectral radius of the Hessian of the loss function. We also derive algorithms to efficiently optimize neural network models and prove that these algorithms almost surely converge. Furthermore, we demonstrate that our algorithm works effectively on applications in different domains, including healthcare. To show that our models generalize well, we introduced various methods for testing generalizability and found that our models outperform comparable baseline models on these tests.
\end{abstract}

\section{Introduction}
\label{Introduction}

Finding flat minima solutions to optimization problems is important, especially in machine learning applications. Such models generalize better than sharp minima because the value of the loss function remains similar around flat minima if the data is shifted, distorted, or otherwise changed. Thus, in practice, optimal machine learning models near flatter optima should perform better than those near sharper minima on test data distributed differently than the original training data \citep{keskar2017on}.

Here, we define flat minima as those with a small spectral radius of the Hessian of the loss function (i.e., the largest absolute eigenvalue is small) and sharp minima as those where the spectral radius is large. For flat minima, there is no direction away from the minimum in which the loss function immediately and rapidly increases or decreases. Therefore, by regularizing the optimization of models with respect to this spectral radius, we can obtain solutions that are less susceptible to errors and biases in training or test data.

However, this regularization presents certain challenges. For large neural networks, computing and storing the Hessian and the third derivative tensor (used in the gradient of the spectral radius term) are intractable; therefore, we develop methods to efficiently compute the regularization term and its gradient without computing these full quantities. We also design methods to introduce errors and biases into the data to test the generalizability of these models.

To tackle these challenges, we build methods to regularize the spectral radius while computing Hessian-vector products, rather than computing the full Hessian and then multiplying by the vector. We approximate the spectral radius and corresponding eigenvector using algorithms such as power iteration and Locally Optimal Block Preconditioned Conjugate Gradient (LOBPCG). We extend methods used for computing Hessian-vector products for neural networks and use them to efficiently compute the eigenvector and spectral radius gradient, used in our algorithms. Implementing these methods within a batch stochastic gradient descent algorithm allows us to optimize a neural network with a given loss function and our regularization term.

We also present results with different regularization parameters to show that our methodology is stable.
Our contributions are as follows.

\begin{itemize}
\item We develop algorithms for regularizing neural networks with respect to the spectral radius of the Hessian, a novel use of a derivative measure for such regularization.
\item We derive differential operators for efficient computation of Hessian-vector products for neural networks.
\item We provide formal proofs of convergence and other properties of our algorithm.
\item We present experimental results on multiple real-world data sets across different domains, designing specific methods to test generalizability.
\end{itemize}

In Section \ref{Literature Review}, we review existing literature related to our research. In Section \ref{Algorithm}, we derive the algorithm used for our regularization. In Section \ref{Analysis}, we discuss convergence results and other properties of the algorithm. In Section \ref{Experiments}, we describe different generalizability tests and present the results of our experiments with regularization on various data sets.

\section{Related Work}
\label{Literature Review}

Existing research discussed how different learning methods affect the ability of neural networks to converge to flat minima. \cite{keskar2017on} observed that large-batch stochastic gradient descent (SGD) and its variants, such as adaptive moment estimation (Adam), tend to converge to sharp minima. In contrast, small-batch methods converge to flat minima. This implies that small-batch methods generalize better than large-batch methods, as the training function at sharp minima is more sensitive. Some possible causes include large-batch methods over-fitting, being attracted to saddle points, and lacking the exploratory properties of small-batch methods (i.e., they tend to converge to the minima close to the initial weights).
\cite{yao2018hessianbased} showed that large-batch training of neural networks converges to points with a larger Hessian spectrum (both in terms of dominant and other eigenvalues), showing poor robustness. 
\cite{jastrzebski2018finding, Zhang2024} extended these claims by showing that a large learning rate also leads to flatter minima that generalize better than sharper minima.
\cite{PhysRevLett.127.278301} showed that wide flat minima in nonconvex neural networks arise as structures from groups of minima around locally robust configurations.
\cite{Wu2022} showed that SGD favors flat minima but left the connection between the Hessian and generalization as an open question for future work.

Others used different ways to measure and find flat minima, including loss functions and optimization algorithms.
\cite{Linjian} suggested that Kronecker-Factored Approximate Curvature (K-FAC) \citep{MartensG15}, an approximate second-order method, may yield generalization improvements over first-order SGD. \cite{chaudhari2017entropysgd, pmlr-v80-dziugaite18a, pittorino2021entropic} proposed an entropy-based loss function to find solutions in flat regions and an algorithm (called entropy-SGD) to optimize models.
\cite{He2019} observed that at local minima of deep networks, there exist many asymmetric directions where the loss sharply increases, which they call "asymmetric valleys." They proposed stochastic weight averaging (SWA) along the SGD trajectory to bias solutions towards the flat side. \cite{chaudhari2017entropysgd} also noted that many neural networks, trained on various data sets using SGD or Adam, converge to a point with a large number of near-zero eigenvalues, along with a long positive tail and shorter negative tail. Our regularization method, which attempts to reduce the spectral radius of the Hessian, is tailored to avoid the eigenspectrum asymmetries described by \cite{chaudhari2017entropysgd} and \cite{He2019}.

\cite{foret2021sharpnessaware} developed a Sharpness-Aware Minimization (SAM) algorithm, which minimizes the maximum loss within a neighborhood of a point. Unlike us, they focused on testing model generalization on fuzzy labels. \cite{andriushchenko2022understanding} have raised doubts about SAM's ability to generalize in other settings. Adding SAM as a baseline comparison would require training on multiple seeds due to the stochasticity. Additionally, the SAM paper was published after the original draft of this paper was posted to Ar{X}iv.

While \cite{yoshida2017spectral} developed a spectral norm radius regularization method, it looks solely at the spectral radius of a neural network's weight matrices rather than the spectral radius of the Hessian of the loss function. Though they experimentally showed that their regularization method has a small generalization gap (between the training and test set), their method also had a higher Hessian spectral radius than vanilla SGD, weight-decay, and adversarial methods. We believe our regularization method and generalization tests more directly address finding flat minima and measuring their generalizability.

\cite{Kaddour2022} compared SWA and SAM in various computer vision, natural language processing, and graph representation learning tasks. They concluded that the effectiveness of these methods is influenced by the dataset and model architecture. Flat-minima optimizers can offer asymmetric payoffs, potentially leading to slight performance decreases at worst, but significant gains at best.

\section{Algorithm}
\label{Algorithm}

\begin{table}[t]
\centering
\begin{tabular}{l|l}
Variable & Definition \\ \hline
$w$ & model parameters or weights \\ \hline
$f(w)$ & loss function \\ \hline
$H(w)$ & Hessian of $f(w)$ \\ \hline
$\rho(w)$ & spectral radius of $H(w)$ \\ \hline
$\mu$ & degree of regularization  \\ \hline
$K$ & goal of $\rho(w)<K$ \\ \hline
$\bar{v}$ & eigenvector corresponding to spectral radius \\
\end{tabular}
\caption{Variable Definitions}
\label{Variables}
\end{table}

We summarize the main variables used and their corresponding definitions in Table \ref{Variables}. We choose to express our problem as a regularized optimization problem rather than a constrained optimization or min-max problem, as strict adherence to our spectral radius constraint is typically unnecessary. Additionally, the regularized approach keeps the algorithm simple, while complexity is computationally taxing for large neural networks. Thus, our optimization problem is
$$\min\limits_w f(w) + \mu \max\{0,\rho(w)-K\},$$
for weights $w\in \R^n$, non-convex loss function $f(w)$, spectral radius (i.e., the maximal absolute eigenvalue) $\rho(w)$ of the Hessian $H(w)$ of $f(w)$, and regularization parameters $\mu$ and $K$. This can also be viewed as Lagrangian relaxation of constraint $\rho(w)\le K$. For convenience, we denote
$$g(w):=f(w) + \mu \max\{0,\rho(w)-K\}.$$

Our goal is to design efficient algorithms for solving this minimization problem, with the caveat that we cannot directly compute $H(w)$. For large neural networks of size $\bigO(n)$, computing and storing objects of size $\bigO(n^2)$ (such as the Hessian) is intractable. However, we can efficiently compute the Hessian-vector product $H(w)v$ for a given $v\in \R^n$ using a method discussed in Section \ref{Hessian-Vector Operations}.

In Section \ref{Algorithms}, we present and explain different variants of our algorithm. In Section \ref{Computing Rho}, we discuss how to compute the regularized term and its gradient.

\subsection{Algorithms}
\label{Algorithms}

Here, we present two versions of our algorithm: a batch stochastic gradient descent power iteration algorithm (Algorithm \ref{SGD}) and a LOBPCG algorithm (Algorithm \ref{LOBPCG}).
The LOBPCG method tries to improve the run time of power iteration by using a preconditioner (a transformation used to improve numerical methods). For simplicity, we hide the $w_k$ dependencies (where $w_k$ is the value of weights $w$ at iteration $k$) for many of the variables by defining: $f_k:=f(w_k)$, $g_k:=g(w_k)$, $\rho_k:=\rho(w_k)$, $\nabla f_k := \nabla f(w_k)$, etc. We let the step size $\alpha_k$ be a predefined function of iteration $k$ and $L$ be the maximum number of iterations. We assume $f(w)=\sum\limits_i \bar{f}^{(i)} (w)$ and write $\bar{f}^{(i)}_k:=\bar{f}^{(i)} (w_k)$ as the value of the loss function $f$ on sample $i$ at iteration $k$. We also let $\bar{H}^{(i)}_k$ be the Hessian matrix of $\bar{f}^{(i)}$ at $w_k$.

\begin{algorithm}[ht]
Initialize $w_1$ \\
\For{$k=1,\cdots,L$}{
Select batch $B_k$ of cardinality $\U$ uniformly at random \\
Compute $\nabla f_k=\frac{1}{\U}\sum\limits_{i\in B_k} \nabla \bar{f}^{(i)}_k$ \\
Initialize $u$, $\lambda$, and $v$ \\
\While{$||u-\lambda v||>\varepsilon_k$}{
$u = \frac{1}{\U}\sum\limits_{i\in B_k}\bar{H}^{(i)}_kv$ (using $\rop{\cdot}$) \\
$\lambda = u^T v$ \\
$v = \frac{u}{||u||}$ \\
}
$\rho_k=\lambda,\ v_k=v$ \\
$\nabla \rho_k=\frac{1}{\U}\sum\limits_{i\in B_k} v_k^T\nabla \bar{H}^{(i)}_k v_k$ (using $\rrop{\cdot}$) \\
Update $p_k = \nabla f_k + \mu \nabla \rho_k \mathbbm{1} \left(\rho_k>K\right)$ \\
$w_{k+1}=w_k-\alpha_k p_k$ \\
}
\caption{Batch Stochastic Gradient Descent}
\label{SGD}
\end{algorithm}

We start with Algorithm \ref{SGD}, a batch stochastic gradient descent algorithm, which uses power iteration to compute $\rho_k$ and $\nabla \rho_k$. 
Due to the implementation of $\rop{\cdot}$ and $\rrop{\cdot}$ (see Section \ref{Computing Rho}) during these computations, the storage requirements of $\bigO(n)$ are not onerous. Also, since the Hessian is symmetric, power iteration converges at a rate proportional to the square of the ratio between the two largest eigenvalues $\bigO\left(|\lambda_1/\lambda_2|^2\right)$, rather than the typical linear rate $\bigO\left(|\lambda_1/\lambda_2|\right)$. Note that the gradient computation in Line 4 can be done as part of the $\rop{\cdot}$ in Lines 7 or 11.

\begin{algorithm}[ht]
Initialize $w_1$ \\
\For{$k=1,\cdots,L$}{
Select batch $B_k$ of cardinality $\U$ uniformly at random \\
\If{$k \mod b = 0$}{
Update K-FAC matrix $T$
}
Compute $\nabla f_k=\frac{1}{\U}\sum\limits_{i\in B_k} \bar{f}^{(i)}_k$ \\
Initialize $r$ and $v$ \\
\While{$||r||>\varepsilon_k$}{
$u = \frac{1}{\U}\sum\limits_{i\in B_k}\bar{H}^{(i)}_k v$ (using $\rop{\cdot}$) \\
$\lambda = u^T v$ \\
$r = u - \lambda v$ \\
$w = v + \tilde{\alpha} T r$ \\
$v = \frac{w}{||w||}$ \\
}
$\rho_k=\lambda,\ v_k=v$ \\
$\nabla \rho_k=\frac{1}{\U}\sum\limits_{i\in B_k} v_k^T\nabla \bar{H}^{(i)}_k v_k$ (using $\rrop{\cdot}$) \\
Update $p_k = \nabla f_k + \mu \nabla \rho_k \mathbbm{1} \left(\rho_k>K\right)$ \\
$w_{k+1}=w_k-\alpha_k p_k$ \\
}
\caption{LOBPCG Method}
\label{LOBPCG}
\end{algorithm}

To improve the run time and convergence of our power iteration method, we developed a LOBPCG method (Algorithm \ref{LOBPCG}). This method uses a step-size $\tilde{\alpha}$ (not necessarily fixed), preconditioner $T$ (for example, K-FAC), and update frequency $b$. The LOBPCG algorithm may converge faster than the power iteration algorithm with good choices for these parameters. \cite{Knyazev} and \cite{Knyazev_2007} assumed and numerically showed that $T$ must be symmetric positive definite, with an efficient preconditioner being an approximation of $H_k^{-1}$ (as the condition number $\kappa (T H_k)$ is low). We chose to use \citepos{MartensG15} K-FAC as the preconditioner, as it satisfies these conditions and is well-suited for neural networks.

\subsection{Gradients of Regularization Term}
\label{Computing Rho}

The spectral radius can be expressed as $\rho(w)=\bar{v}^TH(w)\bar{v}$, where $\bar{v}$ is the eigenvector corresponding to the maximum absolute eigenvalue. To compute gradient update steps for the regularization term, we calculate $\nabla \rho$ using Lemma \ref{Derivative} from \cite{Aa}.

\begin{lemma}
\label{Derivative}
For distinct eigenvalues of a symmetric matrix $A(x): \R \to \R^{n\times n}$,
$$\frac{d \lambda_i(x)}{d x}=\bar{v}_i^T\frac{d A(x)}{d x}\bar{v}_i,$$
where $\bar{v}_i$ is the eigenvector for eigenvalue $\lambda_i$.
\end{lemma}

The expression for this derivative is more complicated with repeating eigenvalues, so we assume that the eigenvalue in question is distinct (in practice, this is usually the case).

Using this result and assumption, we express $\nabla \rho(w)= \bar{v}^T \nabla H(w) \bar{v}$. Thus, by efficiently computing $H(w)v$ and $v^T \nabla H(w) v$ for $w,v\in\R^n$, we can calculate $\rho(w)$ and $\nabla \rho(w)$, respectively.

\subsubsection{Hessian-Vector Operations}
\label{Hessian-Vector Operations}

In order to compute $H(w)v$ and $v^T \nabla H(w) v$ for large neural networks with $w,v\in\R^n$, we extend \citepos{Pearlmutter94fastexact} operator $v \to \ropt{v}{f; w}$, defined as
$$\ropt{v}{f; w}:=\left.\frac{\partial}{\partial r}f(w+rv)\right |_{r=0}.$$
Note that $\ropt{v}{\nabla f; w}=H(w)v$. Thus, by applying the differential operator $\ropt{v}{\cdot}$ to the forward and backward passes used to calculate the gradient, we can compute $\rho(w)$ efficiently.

We extend this operation to
$$\rropt{v}{f; w}:=\ropt{v}{\ropt{\cdot}{f; w};w}$$
by applying the differential operator $\ropt{v}{\cdot}$ again to the forward and backwards passes. Particularly, we compute $\rropt{v}{x}$ and $\rropt{v}{y}$ during the forward pass and $\rropt{v}{\nabla_y f}$, $\rropt{v}{\nabla_x f}$, and $\rropt{v}{\nabla_w f}$ during the backward pass, where $\nabla_y$, $\nabla_x$, and $\nabla_w$ are the gradients with respect to output $y$, input $x$, and weights $w$. We derive our formulas in Appendix \ref{A. Hessian-Vector Operations}. Since $\rropt{v}{\nabla f; w}=v^T \nabla H(w) v$, this allows us to efficiently compute $\nabla \rho(w)$.

These methods keep the number of stored values $\bigO(n)$, while directly computing the Hessian and third derivative tensor would require $\bigO(n^2)$ and $\bigO(n^3)$ storage (which is intractable for large networks).

\section{Algorithm Convergence Analysis}
\label{Analysis}

Here, we show that Algorithms \ref{SGD} and \ref{LOBPCG} almost surely converge to a critical point, with some assumptions. While we outline our proofs here, the details are in Appendix \ref{A. Analysis}.

We assume that batches $B$ are randomly selected. Note that $p_k=p_k(w_k)$. We made the following assumptions.

\begin{enumerate}[wide=0pt,label=\textbf{A\arabic*},ref=A\arabic*]
\item \label{function} $f: \R^n \to \R$, $f\in C^5$, $g(w)$ is bounded from below (without loss of generality, $g(w)\ge0$).

\item \label{learning rate} Conditions on the learning rate and tolerance: $$\sum\limits_{k=1}^\infty \alpha_k^2 < \infty,  \sum\limits_{k=1}^\infty \alpha_k = \infty, \sum\limits_{k=1}^\infty \varepsilon_k \alpha_k < \infty.$$

\item \label{moments} The moments do not grow too quickly:
$$\begin{aligned}\left|\left|\frac{1}{\U}\sum\limits_{i\in B} \nabla \bar{f}^{(i)}(w)\right|\right|^j\le A_j^{(1)} + B_j^{(1)} ||w||^j, \\
\left|\left|\frac{1}{\U}\sum\limits_{i\in B}v^{T}\bar{H}^{(i)}(w) v\right|\right|^j\le A_j^{(2)} + B_j^{(2)} ||w||^j,\end{aligned}$$ \\
for $j=2,3,4$ and all $k$; positive constants $A_j^{(1)}$, $A_j^{(2)}$, $B_j^{(1)}$, $B_j^{(2)}$, any weights $w$ and unit vector $v$, and any subset $B$ of cardinality $\U$.

\item \label{lipschitz} The Hessian $H(w)$ is Lipschitz continuous.

\item \label{epsilonK} We have $\varepsilon_k\le 1$ and $\varepsilon_k\to0$ as $k\to\infty$.

\item \label{horizon} Outside a certain horizon, the gradient points toward the origin. There exists $D<\infty$ such that
$$\inf\limits_{||w||^2\ge D,\ v} w^T\sum\limits_{i\in B}\left[ \nabla \bar{f}^{(i)}(w)+\mu v^{T}\nabla \bar{H}^{(i)}(w) v\right] >0,$$
for any subset $B$ of cardinality $\U$. There are well-known tricks to ensure this assumption, such as adding a small linear term \citep{Bottou98on-linelearning}.
\end{enumerate}

First, the stopping criteria for computing the eigenvector is met.

\begin{lemma}
\label{eigenvector}
Power Iteration (Steps 7-10 in Algorithm \ref{SGD}) and LOBPCG (Steps 9-14 in Algorithm \ref{LOBPCG}) always finish in a finite number of iterations with
$||v_k-\bar{v}_k||\le \varepsilon_k$, where $\bar{v}_k$ is an eigenvector corresponding to the leading eigenvalue of $H_k=\frac{1}{|B|}\sum_{i\in 
B}\bar{H}_k^{(i)}$.
\end{lemma}

This follows from the proofs of power iteration convergence by \cite{Parlett} and LOBPCG convergence by \cite{Knyazev}.

\begin{lemma}
\label{epsilon}
Given Assumptions \ref{function}-\ref{horizon}, $\lim\limits_{k\to\infty}v_k^T\nabla H_k v_k=\lim\limits_{k\to\infty}\nabla \bar{\rho}_k$, where $\nabla \bar{\rho}_k$ is the true gradient of the Hessian's spectral radius.
\end{lemma}

We split $v_k^T\nabla H_k v_k$ into components for the true eigenvector $\bar{v}_k$ and our estimate $v_k$. Then, we bind it, showing that Assumption \ref{epsilonK} is sufficient for Lemma \ref{epsilon} to hold.

Then, we show that these stochastic algorithms fit our bounds on the moments of the update term. Here, we take the expectation with respect to the choice of batch $B_k$, conditioned on the history
$$\mP_k:= B_1,\cdots,B_{k-1},\ w_1,\cdots,w_k,\ \alpha_0,\cdots,\alpha_k.$$

\begin{lemma}
\label{Update}
Given Assumptions \ref{moments}-\ref{lipschitz},
$$\E_{B_k}\left[||p_k||^j\big|\mP_k\right]\le A_j + B_j ||w_k||^j,$$
for $j=2,3,4$, positive constants $A_j$ and $B_j$, and any $k$.
\end{lemma}

We split $p_k$ into its components
$p_k:=\nabla f_k+\mu\nabla \rho_k.$
We use the Assumptions to bind each of these components. Then, we combine the results to show that the lemma holds.

Next, we show that the iterates are confined.

\begin{lemma}
\label{Confinement}
Given Assumptions \ref{function}-\ref{horizon}, the iterates $w_k$ in Algorithms \ref{SGD} and \ref{LOBPCG} are bounded almost surely.
\end{lemma}

We define a sequence that is a function of $w_k$ and show that the sum of its positive expectations is finite. Then, we apply the Quasi-Martingale Convergence Theorem and show that since the sequence converges almost surely, the norm of our weights $w_k$ is bounded. Next, using our assumptions and Lemma \ref{Confinement}, we prove almost sure convergence.

\begin{theorem}
\label{Convergence}
Given Assumptions \ref{function}-\ref{horizon}, in Algorithms \ref{SGD} and \ref{LOBPCG} the loss function values $g(w_k)$ converge almost surely and $\nabla g(w_k)$ converge almost surely to 0.
\end{theorem}

We use confinement of $w_k$ to show that positive expected variations in $g(w)$ between iterates are bounded by a constant times our learning rate squared $\alpha_k^2$. Using Assumption \ref{learning rate} and the Quasi-Martingale Convergence Theorem, we show that $g_k$ converges almost surely. Then, we show that $\nabla g_k$ almost surely converges to zero. Our proofs of Lemma \ref{Confinement} and Theorem \ref{Convergence} are based on \citepos{Bottou98on-linelearning} proof that SGD almost surely converges.

\section{Experiments}
\label{Experiments}

We tested our spectral radius regularization algorithms on the following data sets: forest cover types \citep{cov}, United States Postal Service (USPS) handwritten digits \citep{USPS}, and chest X-rays \citep{chestxray}. The forest cover-type data uses cartographic data to predict which of seven tree species is planted on a plot of land. The USPS digits data includes images of digits 0-9 from scanned envelopes. The chest X-ray data uses images to identify which of the fourteen lung diseases patients were diagnosed with. We further describe these data sets in Appendix \ref{A. Data}.

Additionally, we trained unregularized, \citepos{He2019} asymmetric valley, \citepos{chaudhari2017entropysgd} entropy-SGD, and \citepos{MartensG15} K-FAC models, which serve as baseline comparisons. These other methods for finding flat minima were discussed in Section \ref{Literature Review} and serve as baseline comparisons.

\subsection{Setup}

To test if models with lower spectral radii generalize better than those with higher spectral radii, we created test sets that are differently distributed from the training data. To accomplish this, we employed covariate shifts and image augmentation techniques and introduced new, distinct data. We provide a more detailed description of our software, parameter values, and architectures in Appendices \ref{A. Generalization} and \ref{A. Implementation}.

For the forest cover-type data, we weighted the test plots of land to shift the mean of the features. Then, we compared the accuracy of the trained models and repeated them for one thousand shifts. These perturbations simulate test conditions with poor measurements or climate changes. This weighting method is opposite to \cite{Hidetoshi2000, Huang2006}; we made the test and training data have different, rather than similar, distributions.

For USPS digits, we augmented the test set using random crops and rotations, a subset of the perturbations used by \cite{Hendrycks2019} to benchmark robustness on ImageNet. These modifications simulate test conditions where digits are written on angles, cut off, or poorly scanned. We also compared how models trained on USPS data performed on MNIST \citep{MNIST} and images from Conditional Generative Adversarial Networks (GANs) \citep{CGAN}. \cite{Zhang2021} used performance on GAN-generated data to predict generalizability.

For the chest X-ray models, we compared performance on two similar transfer learning data sets, CheXpert \citep{CheXpert} and MIMIC-CXR \citep{MIMIC-CXR} (using the six conditions common to all three data sets). We kept the labeled training and validation sets separate due to differences in how the conditions were recorded. As the new chest X-ray data contains different patients with conditions not present in the training data, it tests how well these models perform in different populations. \cite{Kim2019DesignCO, Salehinejad2021ARD} stated that the use of data from multiple geographically and temporally distinct sources is important to demonstrate the generalizability of medical image models. \cite{Zech2018VariableGP} showed that a CheXNet model \citep{CheXNet} trained to detect pneumonia generalized poorly to data from other hospital systems and times. Since this is a multi-class, multi-label problem, we measure performance using the mean area under the curve (AUC) of the receiver operating characteristic curve over each class.

We selected different regularization parameters $\mu$ and $K$ via an informal grid search. If $\mu$ is too large, the model will converge to a flat outlying point, predicting the same class for each sample. If $\mu$ is too small, the regularization will be ineffective.

Details on hyperparameters, network architectures, hardware, and software are in Appendix \ref{A. Experiments}. Data sizes are also discussed in Appendix \ref{A. Data}.

\subsection{Results}
\label{Results}

\begin{table}[t]
\centering
\small
\begin{tabular}{l|r|r|r|r}
& & \multicolumn{1}{c|}{Test} & \multicolumn{2}{c}{Relative Shift Acc.} \\ \cline{4-5}
Model & \multicolumn{1}{c|}{$\rho$} & \multicolumn{1}{c|}{Acc.} & \multicolumn{1}{c|}{Mean} & \multicolumn{1}{c}{95\% CI} \\ \hline \hline
Unregularized & 36.58 & \textbf{71.74} & -2.80\% & [-3.38, -2.23] \\ \hline
Asym. Valley & 23.28 & 70.99 & -1.48\% & [-1.96, -1.01] \\ \hline
Entropy-SGD & 6.82 & 69.69 & -1.41\% & [-1.89, -0.92] \\ \hline
K-FAC & 58.55 & 70.83 & -2.29\% & [-2.85, -1.72] \\ \hline \hline
$\mu$=.01, K=1 & \textbf{1.68} & 69.71 & -1.61\% & [-2.10, -1.13] \\ \hline
$\mu$=.01, K=0 & 2.16 & 70.39 & \textbf{-0.96\%} & \textbf{[-1.31, -0.61]} \\ \hline
$\mu$=.005, K=1 & 3.09 & 70.97 & -1.50\% & [-1.97, -1.02] \\ \hline
$\mu$=.001, K=5 & 7.15 & 70.67 & -1.80\% & [-2.28, -1.32] \\ \hline
$\mu$=.001, K=0 & 9.03 & 70.87 & -1.96\% & [-2.44, -1.48] \\ \hline
LOBPCG & 1.99 & 69.49 & -2.87\% & [-3.45, -2.29] \\
\end{tabular}
\caption{Comparison of forest cover-type models. Performance on the shifted test data is measured relative to each model's test accuracy (accuracy on shifted data divided by accuracy on test data minus 1). Optimal values are bolded.}
\label{Forest Cover}
\end{table}

We trained a feed-forward neural network on forest cover-type data and compared the accuracy of models on the randomly shifted test sets. We selected different regularization parameters $\mu$ and $K$ via an informal grid search. Table \ref{Forest Cover} shows a benefit to the asymmetric valley, entropy-SGD, and power iteration regularized models over the unregularized model. The K-FAC and LOBPCG models do not significantly outperform the unregularized model. While there are some differences between the various regularized models -- there is some delineation between those with lower $\rho$ and higher -- all generalize better than the unregularized model. The relative decrease in accuracy on the shift data is less on the regularized models than on the unregularized model. Also, our spectral radius measure $\rho$ mostly follows the regulation strictness. Our strictest regularized model with $\mu=0.01, K=0$ and small $\rho=2.16$ saw the lowest decrease in accuracy on the shifted data. The confidence intervals show that models that performed worse on the shifted data also had a higher variance in their results. We further discuss the LOBPCG results in Section \ref{Spectral Radius Computation}.

\begin{figure*}
    \centering
    \begin{subfigure}[t]{.49\textwidth}
    \centering
    \includegraphics[scale=.38]{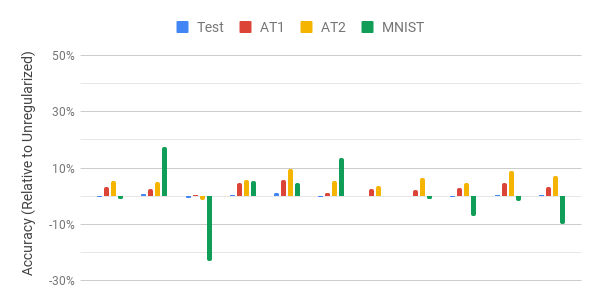}
    \label{USPS1}
    \end{subfigure}
    \begin{subfigure}[t]{.49\textwidth}
    \centering
    \includegraphics[scale=.38]{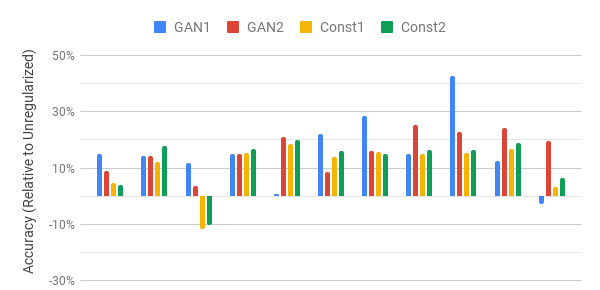}
    \label{USPS2}
    \end{subfigure} \\[-2ex]
    \begin{subfigure}[t]{.49\textwidth}
    \centering
    \includegraphics[scale=.38]{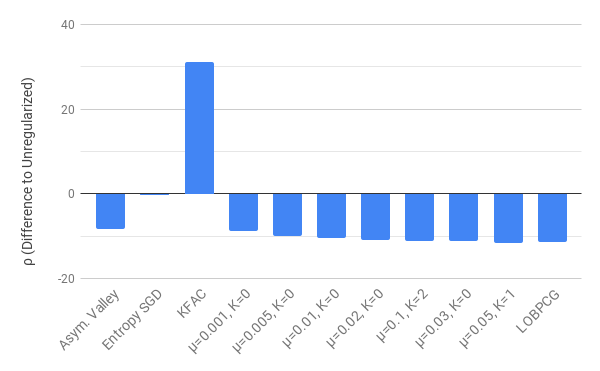}
    \label{USPS Rho}
    \end{subfigure}
    \begin{subfigure}[t]{.49\textwidth}
    \centering
    \includegraphics[scale=.38]{USPS/USPSRho.png}
    \end{subfigure}
\caption{Accuracy of models trained on USPS data. Augmented Test (AT) 1 uses random crops of up to one pixel and random rotations of up to 15$^\circ$. AT 2 uses crops of up to two pixels and rotations of up to 30$^\circ$.}
\label{USPS}
\end{figure*}

\begin{figure}[t]
    \centering
    \includegraphics[scale=0.37]{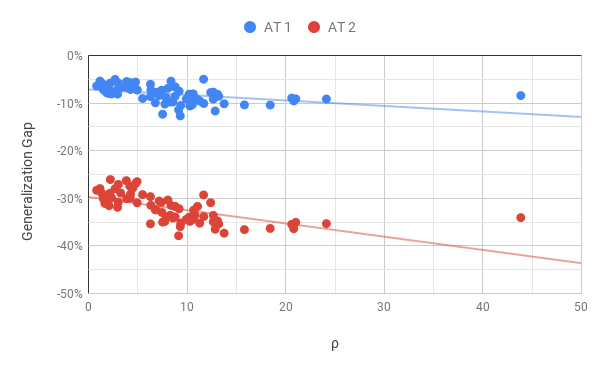}
    \caption{The generalization gap (augmented test accuracy divided by test accuracy minus 1) of USPS models tends to be larger for models with larger spectral radius $\rho$. A dotted linear trend line is given for reference.}
    \label{USPS Gen Gap}
\end{figure}

We trained a convolutional neural network on USPS digits, using various regularization and optimization methods, and compared the accuracy on multiple test sets. Per Figure \ref{USPS}, while the models performed comparably on the test data (all models have an accuracy of 94.47-95.91\%), our regularized models (both power iteration and LOBPCG) performed significantly better than the unregularized model on both augmented test data sets (87.10-91.08\% vs. 86.20\% on Augmented Test 1; 65.37-69.06\% vs. 63.03\% on Augmented Test 2). Our regularized model with $\mu=0.005$ and $K=0$ was the most accurate on the USPS and augmented test sets. The model with the lowest spectral radius (1.16, $\mu=0.05$ and $K=1$) performed second-best on the augmented data. The asymmetric valleys model outperformed the other baseline models but was still 2.2\%-3.9\% worse than the $\mu=0.005$ and $K=0$ model on the augmented data. Also, Figure \ref{USPS} shows that there is a clear relationship between our regularization parameter $\mu$ and the spectral radius $\rho$ for our regularized models: as $\mu$ increases, $\rho$ decreases (provided it is greater than $K$). Figure \ref{USPS Gen Gap} uses 70 models with different spectral radii to show that as $\rho$ increases, the magnitude of the generalization gap (between the test and augmented test sets) increases. This implies that models with higher spectral radii tend to perform worse on this generalization task.

We also trained two GANs on USPS data using \citepos{Linder-Noren} (GAN1) and \citepos{Chhabra} (GAN2) methodology. We found that the images generated by GAN1 were too similar, causing the models to classify or misclassify them in the same way. For example, the $\mu=0.005$ and $K=0$ model misclassified 0's as 2's, 5's as 3's, and 9's as 6's. GAN2 did not suffer from this issue. The figure on the right of Figure \ref{USPS} shows the results. The model that performed best on the augmented tests ($\mu=0.005$ and $K=0$) performed fourth-best on the MNIST (59.80\%) and GAN2 data (88.15\%) but performed poorly on GAN1 (70.44\%) due to the aforementioned issues. The regularized models with $\rho\approx1.51$ performed best on the GAN data; the regularized model with $\mu=0.03$ and $K=0$ was 99.82\% accurate on GAN1, and the $\mu=0.1$ and $K=2$ model was 91.35\% accurate on GAN2. Entropy-SGD performed best on the MNIST data (67.12\%), but regularized models were the next four best-performing models. While examining the GAN results, we realized that the generated images were abnormally distributed relative to the USPS images. To determine if these results were coincidental or due to this distribution, we constructed two data sets, Const1 and Const2, from the augmented test data to mimic the abnormal image distribution in GAN1 (see Appendix \ref{Constructed}) and found that the $\mu=0.005$ and $K=0$ model performed best on the constructed data. Thus, we conclude that the GAN1 results appear to be a coincidence and recommend the $\mu=0.005$ and $K=0$ model.

\begin{figure}[t]
    \centering
    \begin{subfigure}[t]{.48\textwidth}
    \centering
    \includegraphics[scale=.37]{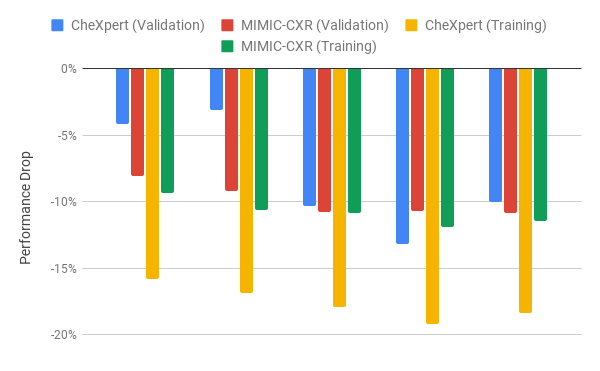}
    \label{Chest X-Ray Performane Drop}
    \end{subfigure}  \\[-4ex]
    \begin{subfigure}[t]{.48\textwidth}
    \centering
    \includegraphics[scale=.37]{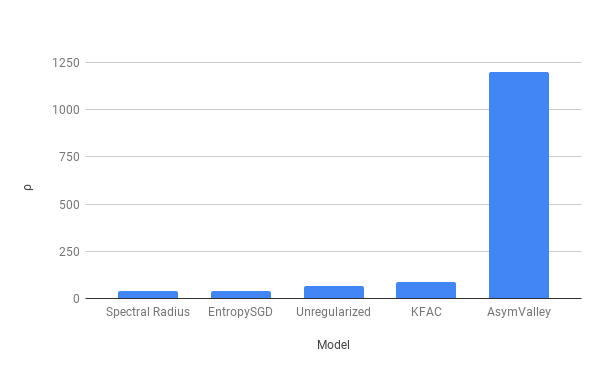}
    \label{Chest X-Ray Spectral Radius}
    \end{subfigure}
\caption{Chest X-ray models with low spectral radius have a lower drop in performance on distinct chest X-ray data. The performance drop was measured as the difference in mean AUC of the 6 overlapping classes from the held-out test data to the transfer learning data.}
\label{Chest X-Ray}
\end{figure}

For chest X-ray comparisons, we trained CheXNet (a 121-layer DenseNet trained on chest X-ray data, based on \href{https://github.com/zoogzog/chexnet}{github.com/zoogzog/chexnet}) as our baseline. Using this model as an initialization, we trained for an additional epoch with our spectral radius regularization method, comparing the mean AUC. Similarly, we used this initialization to train the entropy-SGD, K-FAC, and asymmetric valley models. For our regularized model, we employed gradient clipping to curtail an exploding spectral radius gradient. Figure \ref{Chest X-Ray} shows that the two models with the lowest spectral radius $\rho$, our regularized model ($\mu=10^{-4}$ and $\alpha=10^{-6}$) and entropy-SGD, performed best on the transfer learning chest X-ray data sets. Our model had a 5.34\% lower $\rho$ than entropy-SGD and a lower performance drop on 3-of-the-4 transfer learning data sets. In Appendix \ref{Grad-CAM}, we use Grad-CAM to highlight the regions of the X-rays used to make predictions. We show that the two models with the lowest spectral radius overlap the most in explanations, signifying that their explanations generalize better too.

\subsection{Batch Size}
\label{batch size}

\begin{table}[t]
\centering
\begin{tabular}{l|r|r|r|r}
Batch & & \multicolumn{1}{c|}{Test} & \multicolumn{2}{c}{Relative Shift Acc.} \\ \cline{4-5}
Size & \multicolumn{1}{c|}{$\rho$} & \multicolumn{1}{c|}{Acc.} & \multicolumn{1}{c|}{Mean} & \multicolumn{1}{c}{95\% CI} \\ \hline \hline
32 & 5.25 & 69.39 & -2.88\% & [-3.46, -2.29] \\ \hline
64 & 4.05 & 68.84 & -3.10\% & [-3.69, -2.52] \\ \hline
128 & 2.16 & \textbf{70.39} & \textbf{-0.96\%} & \textbf{[-1.31, -0.61]} \\ \hline
256 & 1.32 & 69.30 & -1.52\% & [-2.01, -1.04] \\ \hline
512 & \textbf{1.24} & 69.16 & -1.49\% & [-1.98, -1.00] \\ \hline
\end{tabular}
\caption{Effect of batch size on accuracy of forest cover-type models with $\mu=0.01$ and $K=0$.}
\label{Forest Batch}
\end{table}

\begin{table}[t]
\centering
\begin{tabular}{l|r}
Batch Size & \multicolumn{1}{c}{$\rho$} \\ \hline \hline
32 & 4.82 \\ \hline
64 & 3.31 \\ \hline
128 & 2.16 \\ \hline
256 & 1.39 \\ \hline
512 & 0.93 \\
\end{tabular}
\caption{Computed spectral radius of forest cover-type model with $\mu=0.01$ and $K=0$ trained with batch size of 128.}
\label{Forest Batch 2}
\end{table}

\begin{table*}[t]
\centering
\begin{tabular}{l|r|r|r|r|r|r|r|r|r}
Batch & & \multicolumn{8}{c}{Accuracy} \\ \cline{3-10}
Size & \multicolumn{1}{c|}{$\rho$} & Test & AT 1 & AT 2 & MNIST & GAN1 & GAN2 & Const1 & Const2 \\ \hline \hline
32 & 7.88 & 93.27 & 85.00 & 61.58 & 55.21 & \textbf{90.07} & 86.18. & 70.47 & 71.67 \\ \hline
64 & 4.93 & 94.47 & 87.64 & 65.58 & 59.30 & 80.77 & 86.52 & 74.24 & 73.54 \\ \hline
128 & 2.69 & \textbf{95.91} & \textbf{91.08} & \textbf{69.06} & \textbf{59.80} & 70.44 & 88.15 & \textbf{78.20} & \textbf{78.54} \\ \hline
256 & 2.14 & 94.42 & 88.59 & 67.76 & 57.99 & 73.28 & 82.31 & 75.87 & 76.04 \\ \hline
512 & \textbf{2.08} & 94.97 & 86.85 & 64.62 & 50.79 & 73.16 & \textbf{89.63} & 72.27 & 73.12 \\ \hline
\end{tabular}
\caption{Effect of batch size on accuracy of USPS models with $\mu=0.005$ and $K=0$.}
\label{USPS Batch}
\end{table*}

As discussed in Section \ref{Literature Review}, \cite{keskar2017on, yao2018hessianbased, jastrzebski2018finding} showed that large-batch training methods yield more generalizable models. This motivated us to analyze the effect of batch size on the spectral radius and generalizability of our regularized forest cover type and USPS models. Contrary to their findings, we found (Tables \ref{Forest Batch} and \ref{USPS Batch}) that smaller batch models have a larger spectral radius. Despite this, the model with the original batch size (128) generally performed best on our comparison tests. However, Table \ref{Forest Batch 2} shows that the batch size is a major contributing factor to the computed spectral radius.

\subsection{Computational Time Breakdown}
\label{Computational Times}

\begin{figure*}
    \centering
    \begin{subfigure}[t]{.49\textwidth}
    \centering
    \includegraphics[scale=.34]{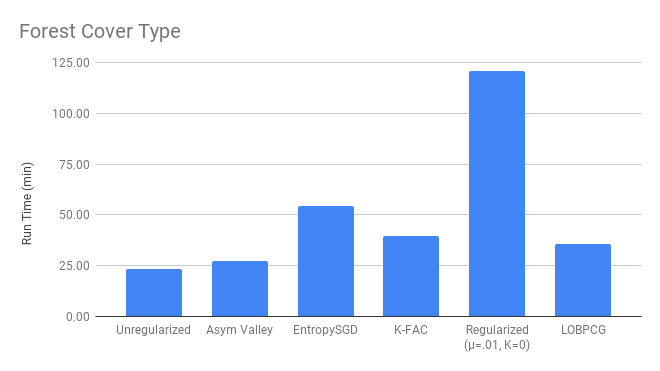}
    \label{Cov_RT}
    \end{subfigure}
    \begin{subfigure}[t]{.49\textwidth}
    \centering
    \includegraphics[scale=.34]{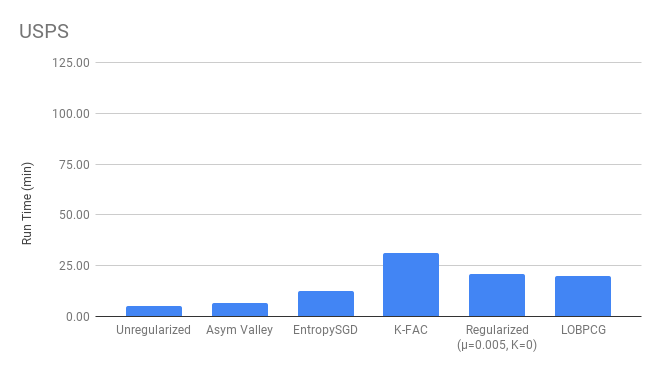}
    \label{USPS_RT}
    \end{subfigure}
\caption{Computational time of models' training on forest cover-type and USPS data.}
\label{Run Time}
\end{figure*}

Figure \ref{Run Time} shows that the computational time of the power iteration regularization method was relatively high but not prohibitively so. The unregularized models were the fastest to train, followed by the asymmetric valley models. The power iteration model took the longest to train on forest cover-type data and the second longest to train on USPS. LOBPCG significantly improved the training time of the forest cover-type model to the point where it was the third-fastest model. However, it only decreased the training time of the USPS model by 3\%. K-FAC took the longest to train on USPS data and the third longest on forest cover type. Entropy-SGD was the second longest on forest cover-type data and the third fastest on USPS.

\begin{figure*}
    \centering
    \begin{subfigure}[t]{.49\textwidth}
    \centering
    \includegraphics[scale=.34]{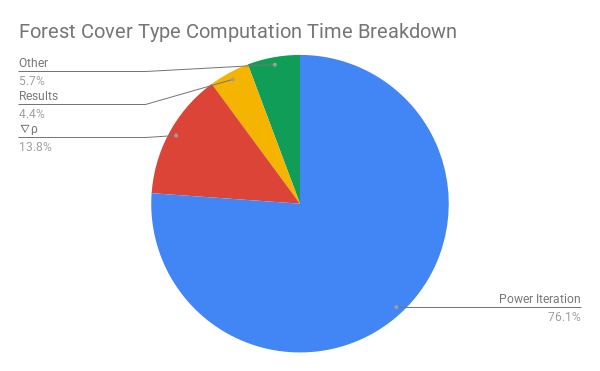}
    \label{Cov_CT}
    \end{subfigure}
    \begin{subfigure}[t]{.49\textwidth}
    \centering
    \includegraphics[scale=.34]{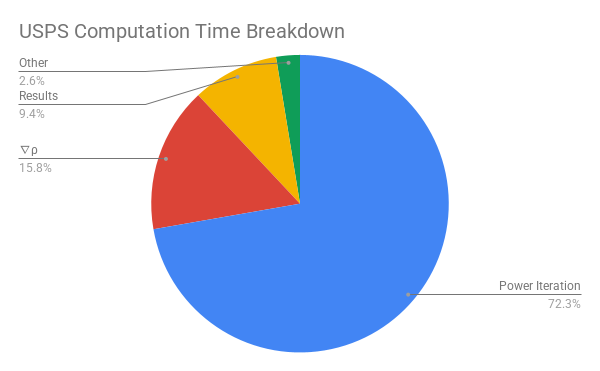}
    \label{USPS_CT}
    \end{subfigure}
\caption{Breakdown of the Algorithm \ref{SGD}'s computational time on forest cover-type and USPS data.}
\label{Time Breakdown}
\end{figure*}

Figure \ref{Time Breakdown} shows that three-quarters of Algorithm \ref{SGD}'s run time is spent on power iteration (lines 5-10). About another 15\% is spent on computing $\nabla \rho_k$ using $\rrop{\cdot}$ (line 11). 5-10\% was spent on computing the results and other tracked statistics.

\subsection{Spectral Radius Computation}
\label{Spectral Radius Computation}

As shown in Section \ref{Computational Times}, the LOBPCG can improve computational training time. These models also had the second-lowest spectral radius (see Section \ref{Results}). However, given the following drawbacks, we cannot universally recommend using this method. It requires additional parameter tuning of the update frequencies $b$ and step sizes $\tilde{\alpha}$. We could not reasonably train a LOBPCG model on the chest X-ray data since the additional memory constraints required a reduced batch size, making the run time onerous. Furthermore, we found that the residual norm ($\left|\left|Hv-\rho v\right|\right|$) was significantly higher for the LOBPCG method than for the power iteration method. Thus, our experiments indicate that the power iteration method generally outperformed the LOBPCG method.

\section{Conclusion}
\label{Conclusion}

We developed algorithms for regularized optimization of neural networks, targeted at finding flat minima. Furthermore, we developed tools for calculating the regularization term and its gradient. We proved that these methods almost surely converge to a critical point. Then, we demonstrated that our regularization generalizes better than baseline comparisons on a range of applicable problems by designing unique methods.

However, optimal performance requires tuning the regularization parameters $\mu$ and $K$ to balance the loss and spectral radius terms, a data- and model-dependent process. We observed that stricter regularization performed better until the regularization was too strict and the model would choose the majority class for all samples. Finding this point requires trial and error.

\section*{Acknowledgment}

Research reported in this publication was supported, in part, by the National Library of Medicine, Grant Number T32LM012203. The content is solely the responsibility of the authors and does not necessarily represent the official views of the National Institutes of Health.

\bibliography{bib}

\begin{thebibliography}{}

\bibitem[\protect\citeauthoryear{Andriushchenko and
  Flammarion}{2022}]{andriushchenko2022understanding}
Maksym Andriushchenko and Nicolas Flammarion.
\newblock Towards understanding sharpness-aware minimization.
\newblock In Kamalika Chaudhuri, Stefanie Jegelka, Le~Song, Csaba Szepesvari,
  Gang Niu, and Sivan Sabato, editors, {\em Proceedings of the 39th
  International Conference on Machine Learning}, volume 162 of {\em Proceedings
  of Machine Learning Research}, pages 639--668. PMLR, 17--23 Jul 2022.

\bibitem[\protect\citeauthoryear{Baldassi \bgroup \em et al.\egroup
  }{2021}]{PhysRevLett.127.278301}
Carlo Baldassi, Clarissa Lauditi, Enrico~M. Malatesta, Gabriele Perugini, and
  Riccardo Zecchina.
\newblock Unveiling the structure of wide flat minima in neural networks.
\newblock {\em Physical Review Letters}, 127:278301, Dec 2021.

\bibitem[\protect\citeauthoryear{Blackard and Dean}{1999}]{cov}
Jock~A. Blackard and Denis~J. Dean.
\newblock Comparative accuracies of artificial neural networks and discriminant
  analysis in predicting forest cover types from cartographic variables.
\newblock {\em Computers and Electronics in Agriculture}, vol.24:131--151,
  1999.

\bibitem[\protect\citeauthoryear{Bottou}{1998}]{Bottou98on-linelearning}
Léon Bottou.
\newblock Online learning and stochastic approximations.
\newblock In {\em Online Learning in Neural Networks}, pages 9--42. Cambridge
  University Press, 1998.

\bibitem[\protect\citeauthoryear{Chaudhari \bgroup \em et al.\egroup
  }{2017}]{chaudhari2017entropysgd}
Pratik Chaudhari, Anna Choromanska, Stefano Soatto, Yann LeCun, Carlo Baldassi,
  Christian Borgs, Jennifer Chayes, Levent Sagun, and Riccardo Zecchina.
\newblock Entropy-{SGD}: Biasing gradient descent into wide valleys.
\newblock In {\em International Conference on Learning Representations}, 2017.

\bibitem[\protect\citeauthoryear{Chhabra}{2021}]{Chhabra}
Sachin Chhabra.
\newblock Py{T}orch-c{GAN}-conditional-{GAN}.
\newblock
  \href{https://github.com/sachin-chhabra/Pytorch-cGAN-conditional-GAN}{github.com/sachin-chhabra/Pytorch-cGAN-conditional-GAN},
  2021.

\bibitem[\protect\citeauthoryear{Dziugaite and
  Roy}{2018}]{pmlr-v80-dziugaite18a}
Gintare~Karolina Dziugaite and Daniel Roy.
\newblock Entropy-{SGD} optimizes the prior of a {PAC}-{B}ayes bound:
  Generalization properties of entropy-{SGD} and data-dependent priors.
\newblock In Jennifer Dy and Andreas Krause, editors, {\em Proceedings of the
  35th International Conference on Machine Learning}, volume~80 of {\em
  Proceedings of Machine Learning Research}, pages 1377--1386. PMLR, 10--15 Jul
  2018.

\bibitem[\protect\citeauthoryear{Foret \bgroup \em et al.\egroup
  }{2021}]{foret2021sharpnessaware}
Pierre Foret, Ariel Kleiner, Hossein Mobahi, and Behnam Neyshabur.
\newblock Sharpness-aware minimization for efficiently improving
  generalization.
\newblock In {\em International Conference on Learning Representations}, 2021.

\bibitem[\protect\citeauthoryear{Gildenblat}{2021}]{jacobgilpytorchcam}
Jacob Gildenblat.
\newblock Py{T}orch library for {CAM} methods.
\newblock
  \href{https://github.com/jacobgil/pytorch-grad-cam}{github.com/jacobgil/pytorch-grad-cam},
  2021.

\bibitem[\protect\citeauthoryear{He \bgroup \em et al.\egroup }{2019}]{He2019}
Haowei He, Gao Huang, and Yang Yuan.
\newblock Asymmetric valleys: Beyond sharp and flat local minima.
\newblock In H.~Wallach, H.~Larochelle, A.~Beygelzimer, F.~d'Alch\'{e} Buc,
  E.~Fox, and R.~Garnett, editors, {\em Advances in Neural Information
  Processing Systems}, volume~32, pages 2553--2564. Curran Associates, Inc.,
  2019.

\bibitem[\protect\citeauthoryear{Hendrycks and
  Dietterich}{2019}]{Hendrycks2019}
Dan Hendrycks and Thomas Dietterich.
\newblock Benchmarking neural network robustness to common corruptions and
  perturbations.
\newblock In {\em International Conference on Learning Representations}, 2019.

\bibitem[\protect\citeauthoryear{Huang \bgroup \em et al.\egroup
  }{2006}]{Huang2006}
Jiayuan Huang, Arthur Gretton, Karsten Borgwardt, Bernhard Sch\"{o}lkopf, and
  Alex Smola.
\newblock Correcting sample selection bias by unlabeled data.
\newblock In B.~Sch\"{o}lkopf, J.~Platt, and T.~Hoffman, editors, {\em Advances
  in Neural Information Processing Systems}, volume~19. MIT Press, 2006.

\bibitem[\protect\citeauthoryear{Irvin \bgroup \em et al.\egroup
  }{2019}]{CheXpert}
Jeremy Irvin, Pranav Rajpurkar, Michael Ko, Yifan Yu, Silviana Ciurea{-}Ilcus,
  Chris Chute, Henrik Marklund, Behzad Haghgoo, Robyn~L. Ball, Katie~S.
  Shpanskaya, Jayne Seekins, David~A. Mong, Safwan~S. Halabi, Jesse~K.
  Sandberg, Ricky Jones, David~B. Larson, Curtis~P. Langlotz, Bhavik~N. Patel,
  Matthew~P. Lungren, and Andrew~Y. Ng.
\newblock Che{X}pert: {A} large chest radiograph dataset with uncertainty
  labels and expert comparison.
\newblock AAAI'19/IAAI'19/EAAI'19. AAAI Press, 2019.

\bibitem[\protect\citeauthoryear{Jastrzebski \bgroup \em et al.\egroup
  }{2018}]{jastrzebski2018finding}
Stanis{\l}aw Jastrzebski, Zachary Kenton, Devansh Arpit, Nicolas Ballas, Asja
  Fischer, Yoshua Bengio, and Amos Storkey.
\newblock Finding flatter minima with {SGD}, 2018.

\bibitem[\protect\citeauthoryear{Johnson \bgroup \em et al.\egroup
  }{2019}]{MIMIC-CXR}
Alistair E.~W. Johnson, Tom~J. Pollard, Seth~J. Berkowitz, Nathaniel~R.
  Greenbaum, Matthew~P. Lungren, Chih{-}ying Deng, Roger~G. Mark, and Steven
  Horng.
\newblock {MIMIC-CXR}, a de-identified publicly available database of chest
  radiographs with free-text reports.
\newblock {\em Scientific Data}, 6(1):317, Dec 2019.

\bibitem[\protect\citeauthoryear{Kaddour \bgroup \em et al.\egroup
  }{2022}]{Kaddour2022}
Jean Kaddour, Linqing Liu, Ricardo Silva, and Matt~J Kusner.
\newblock When do flat minima optimizers work?
\newblock In S.~Koyejo, S.~Mohamed, A.~Agarwal, D.~Belgrave, K.~Cho, and A.~Oh,
  editors, {\em Advances in Neural Information Processing Systems}, volume~35,
  pages 16577--16595. Curran Associates, Inc., 2022.

\bibitem[\protect\citeauthoryear{Keskar \bgroup \em et al.\egroup
  }{2017}]{keskar2017on}
Nitish~Shirish Keskar, Dheevatsa Mudigere, Jorge Nocedal, Mikhail Smelyanskiy,
  and Ping Tak~Peter Tang.
\newblock On large-batch training for deep learning: Generalization gap and
  sharp minima.
\newblock In {\em International Conference on Learning Representations}, 2017.

\bibitem[\protect\citeauthoryear{Kim \bgroup \em et al.\egroup
  }{2019}]{Kim2019DesignCO}
Dong~Wook Kim, Hye~Young Jang, Kyung~Won Kim, Youngbin Shin, and Seong~Ho Park.
\newblock Design characteristics of studies reporting the performance of
  artificial intelligence algorithms for diagnostic analysis of medical images:
  Results from recently published papers.
\newblock {\em Korean Journal of Radiology}, 20:405 -- 410, 2019.

\bibitem[\protect\citeauthoryear{Knyazev \bgroup \em et al.\egroup
  }{2007}]{Knyazev_2007}
Andrew~V. Knyazev, Merico~E. Argentati, Ilya Lashuk, and Evgueni~E.
  Ovtchinnikov.
\newblock Block locally optimal preconditioned eigenvalue xolvers ({BLOPEX}) in
  {H}ypre and {PETSc}.
\newblock {\em SIAM Journal on Scientific Computing}, 29(5):2224–2239, Jan
  2007.

\bibitem[\protect\citeauthoryear{Knyazev}{2001}]{Knyazev}
Andrew~V. Knyazev.
\newblock Toward the optimal preconditioned eigensolver: Locally optimal block
  preconditioned conjugate gradient method.
\newblock {\em SIAM Journal on Scientific Computing}, 23(2), 2001.

\bibitem[\protect\citeauthoryear{LeCun and Cortes}{2010}]{MNIST}
Yann LeCun and Corinna Cortes.
\newblock {MNIST} handwritten digit database.
\newblock 2010.

\bibitem[\protect\citeauthoryear{LeCun \bgroup \em et al.\egroup }{1990}]{USPS}
Yann LeCun, O.~Matan, B.~Boser, {J. S.} Denker, D.~Henderson, {R. E.} Howard,
  W.~Hubbard, {L. D.} Jackel, and {H. S.} Baird.
\newblock Handwritten zip code recognition with multilayer networks.
\newblock In {\em Proceedings - International Conference on Pattern
  Recognition}, volume~2, pages 35--40. Publ by IEEE, 1990.

\bibitem[\protect\citeauthoryear{Linder-Norén}{2021}]{Linder-Noren}
Erik Linder-Norén.
\newblock Py{T}orch generative adversarial networks.
\newblock
  \href{https://github.com/eriklindernoren/PyTorch-GAN}{github.com/eriklindernoren/PyTorch-GAN},
  2021.

\bibitem[\protect\citeauthoryear{Ma \bgroup \em et al.\egroup }{2020}]{Linjian}
Linjian Ma, Gabriel Montague, Jiayu Ye, Zhewei Yao, Asghar Gholami, Kurt
  Keutzer, and Michael Mahoney.
\newblock Inefficiency of k-fac for large batch size training.
\newblock {\em Proceedings of the AAAI Conference on Artificial Intelligence},
  34:5053--5060, 04 2020.

\bibitem[\protect\citeauthoryear{Marcel and Rodriguez}{2010}]{torchvision}
S\'{e}bastien Marcel and Yann Rodriguez.
\newblock Torchvision the machine-vision package of torch.
\newblock In {\em Proceedings of the 18th ACM International Conference on
  Multimedia}, MM '10, page 1485–1488, New York, NY, USA, 2010. Association
  for Computing Machinery.

\bibitem[\protect\citeauthoryear{Martens and Grosse}{2015}]{MartensG15}
James Martens and Roger Grosse.
\newblock Optimizing neural networks with kronecker-factored approximate
  curvature.
\newblock In Francis Bach and David Blei, editors, {\em Proceedings of the 32nd
  International Conference on Machine Learning}, volume~37 of {\em Proceedings
  of Machine Learning Research}, pages 2408--2417, Lille, France, 07--09 Jul
  2015. PMLR.

\bibitem[\protect\citeauthoryear{Mirza and Osindero}{2014}]{CGAN}
Mehdi Mirza and Simon Osindero.
\newblock Conditional generative adversarial nets, 2014.

\bibitem[\protect\citeauthoryear{Parlett and Poole}{1973}]{Parlett}
B.~N. Parlett and W.~G. Poole, Jr.
\newblock A geometric theory for the {QR}, {LU} and power iterations.
\newblock {\em SIAM Journal on Numerical Analysis}, 10(2):389--412, 1973.

\bibitem[\protect\citeauthoryear{Paszke \bgroup \em et al.\egroup
  }{2019}]{pytorch}
Adam Paszke, Sam Gross, Francisco Massa, Adam Lerer, James Bradbury, Gregory
  Chanan, Trevor Killeen, Zeming Lin, Natalia Gimelshein, Luca Antiga, Alban
  Desmaison, Andreas Kopf, Edward Yang, Zachary DeVito, Martin Raison, Alykhan
  Tejani, Sasank Chilamkurthy, Benoit Steiner, Lu~Fang, Junjie Bai, and Soumith
  Chintala.
\newblock Py{T}orch: An imperative style, high-performance deep learning
  library.
\newblock In {\em Advances in Neural Information Processing Systems 32}, pages
  8024--8035. Curran Associates, Inc., 2019.

\bibitem[\protect\citeauthoryear{Pearlmutter}{1994}]{Pearlmutter94fastexact}
Barak~A. Pearlmutter.
\newblock Fast exact multiplication by the {H}essian.
\newblock {\em Neural Computation}, 6:147--160, 1994.

\bibitem[\protect\citeauthoryear{Pittorino \bgroup \em et al.\egroup
  }{2021}]{pittorino2021entropic}
Fabrizio Pittorino, Carlo Lucibello, Christoph Feinauer, Gabriele Perugini,
  Carlo Baldassi, Elizaveta Demyanenko, and Riccardo Zecchina.
\newblock Entropic gradient descent algorithms and wide flat minima.
\newblock In {\em International Conference on Learning Representations}, 2021.

\bibitem[\protect\citeauthoryear{Rajpurkar \bgroup \em et al.\egroup
  }{2017}]{CheXNet}
Pranav Rajpurkar, Jeremy Irvin, Kaylie Zhu, Brandon Yang, Hershel Mehta, Tony
  Duan, Daisy~Yi Ding, Aarti Bagul, Curtis Langlotz, Katie~S. Shpanskaya,
  Matthew~P. Lungren, and Andrew~Y. Ng.
\newblock Che{XN}et: Radiologist-level pneumonia detection on chest {X}-rays
  with deep learning.
\newblock {\em CoRR}, abs/1711.05225, 2017.

\bibitem[\protect\citeauthoryear{Salehinejad \bgroup \em et al.\egroup
  }{2021}]{Salehinejad2021ARD}
Hojjat Salehinejad, Jumpei Kitamura, Noah~G. Ditkofsky, Amy~Wei Lin, Aditya
  Bharatha, Suradech Suthiphosuwan, Hui-Ming Lin, Jefferson~R. Wilson, Muhammad
  Mamdani, and Errol Colak.
\newblock A real-world demonstration of machine learning generalizability in
  the detection of intracranial hemorrhage on head computerized tomography.
\newblock {\em Scientific Reports}, 11, 2021.

\bibitem[\protect\citeauthoryear{Shimodaira}{2000}]{Hidetoshi2000}
Hidetoshi Shimodaira.
\newblock Improving predictive inference under covariate shift by weighting the
  log-likelihood function.
\newblock {\em Journal of Statistical Planning and Inference}, 90:227--244, 10
  2000.

\bibitem[\protect\citeauthoryear{{Van der Aa} \bgroup \em et al.\egroup
  }{2007}]{Aa}
Nico~P. {Van der Aa}, H.G. ter Morsche, and R.M.M. Mattheij.
\newblock Computation of eigenvalue and eigenvector derivatives for a general
  complex-valued eigensystem.
\newblock {\em Electronic Journal of Linear Algebra}, 16:300--314, 2007.

\bibitem[\protect\citeauthoryear{Wang \bgroup \em et al.\egroup
  }{2017}]{chestxray}
Xiaosong Wang, Yifan Peng, Le~Lu, Zhiyong Lu, Mohammadhadi Bagheri, and Ronald
  Summers.
\newblock Chest{X}-ray8: Hospital-scale chest {X}-ray database and benchmarks
  on weakly-supervised classification and localization of common thorax
  diseases.
\newblock In {\em 2017 IEEE Conference on Computer Vision and Pattern
  Recognition (CVPR)}, pages 3462--3471, 2017.

\bibitem[\protect\citeauthoryear{Wang}{2019}]{alecwangcq}
Chaoqi Wang.
\newblock {KFAC}-{P}y{T}orch.
\newblock
  \href{https://github.com/alecwangcq/KFAC-Pytorch}{github.com/alecwangcq/KFAC-Pytorch},
  2019.

\bibitem[\protect\citeauthoryear{Wu \bgroup \em et al.\egroup }{2022}]{Wu2022}
Lei Wu, Mingze Wang, and Weijie Su.
\newblock The alignment property of sgd noise and how it helps select flat
  minima: A stability analysis.
\newblock In S.~Koyejo, S.~Mohamed, A.~Agarwal, D.~Belgrave, K.~Cho, and A.~Oh,
  editors, {\em Advances in Neural Information Processing Systems}, volume~35,
  pages 4680--4693. Curran Associates, Inc., 2022.

\bibitem[\protect\citeauthoryear{Yao \bgroup \em et al.\egroup
  }{2018}]{yao2018hessianbased}
Zhewei Yao, Amir Gholami, Kurt Keutzer, and Michael~W. Mahoney.
\newblock Hessian-based analysis of large batch training and robustness to
  adversaries.
\newblock In {\em Proceedings of the 32nd International Conference on Neural
  Information Processing Systems}, NIPS'18, page 4954–4964, Red Hook, NY,
  USA, 2018. Curran Associates Inc.

\bibitem[\protect\citeauthoryear{Yoshida and
  Miyato}{2017}]{yoshida2017spectral}
Yuichi Yoshida and Takeru Miyato.
\newblock Spectral norm regularization for improving the generalizability of
  deep learning, 2017.

\bibitem[\protect\citeauthoryear{Zech \bgroup \em et al.\egroup
  }{2018}]{Zech2018VariableGP}
John~R. Zech, Marcus~A. Badgeley, Manway Liu, Anthony~Beardsworth Costa,
  Joseph~J. Titano, and Eric~Karl Oermann.
\newblock Variable generalization performance of a deep learning model to
  detect pneumonia in chest radiographs: A cross-sectional study.
\newblock {\em PLoS Medicine}, 15, 2018.

\bibitem[\protect\citeauthoryear{Zhang \bgroup \em et al.\egroup
  }{2022}]{Zhang2021}
Yi~Zhang, Arushi Gupta, Nikunj Saunshi, and Sanjeev Arora.
\newblock On predicting generalization using {GAN}s.
\newblock In {\em International Conference on Learning Representations}, 2022.

\bibitem[\protect\citeauthoryear{Zhang \bgroup \em et al.\egroup
  }{2024}]{Zhang2024}
Jian Zhang, Lei Qi, Yinghuan Shi, and Yang Gao.
\newblock Exploring flat minima for domain generalization with large learning
  rates.
\newblock {\em IEEE Transactions on Knowledge and Data Engineering},
  36(11):6145--6158, 2024.

\end{thebibliography}

\newpage
\appendix
\appendixpage
\section{Hessian-Vector Operations Derivation}
\label{A. Hessian-Vector Operations}

The forward computation for each layer of a network with input $x$, output $y$, weights $w$, activation $\sigma$, bias $I$, error or loss measure $E=E(y)$, and direct derivative $e_k=dE/dy_k$ is given by:
$$\begin{aligned} x_k = x_k(y_{k-1}, w_k) &= \sum\limits_j w_{jk}y_{j(k-1)} \\
y_k = y_k(x_k, I_k) &= \sigma_k(x_k)+I_k\end{aligned}$$
The backward computation:
$$\begin{aligned} \frac{\partial E}{\partial y_k} &= e_k(y_k)+\sum\limits_j w_{jk}\frac{\partial E}{\partial x_j} \\
\frac{\partial E}{\partial x_k} &=\sigma_k'(x_k)\frac{\partial E}{\partial y_k} \\
\frac{\partial E}{\partial w_{jk}} &= y_k\frac{\partial E}{\partial x_j}\end{aligned}$$
Applying $\ropt{v}{\cdot}$ to forward pass:
$$\begin{aligned} \ropt{v}{x_k; w} &= \sum\limits_j \left(w_{jk}\ropt{v}{y_{j(k-1)}; w}+v_{jk}y_{j(k-1)}\right) \\
\ropt{v}{y_k; w} &= \ropt{v}{x_k; w}\sigma_k'(x_k)\end{aligned}$$
The backward computation follows as:
$$\begin{aligned} \ropt{v}{\frac{\partial E}{\partial y_k}; w}=& e_k'(y_k)\ropt{v}{y_k; w} \\
&+\sum\limits_j \bigg[w_{jk}\ropt{v}{\frac{\partial E}{\partial x_j}; w}+v_{jk}\frac{\partial E}{\partial x_j}\bigg] \\
\ropt{v}{\frac{\partial E}{\partial x_k}; w} =&\sigma_k'(x_k)\ropt{v}{\frac{\partial E}{\partial y_k}; w} \\
&+\ropt{v}{x_k; w}\sigma_k''(x_k)\frac{\partial E}{\partial y_k} \\
\ropt{v}{\frac{\partial E}{\partial w_{jk}}; w} =& y_k\ropt{v}{\frac{\partial E}{\partial x_j}; w}+\ropt{v}{y_k; w}\frac{\partial E}{\partial x_j}\end{aligned}$$
This yields the result found in \cite{Pearlmutter94fastexact}. However, we extend it one step further by applying $\ropt{v}{\cdot}$ again, i.e., applying $\rropt{v}{\cdot}=\ropt{v}{\ropt{v}{\cdot}}$ to the original forward pass:
$$\begin{aligned} \rropt{v}{x_k; w} =& \sum\limits_j \Bigl[w_{ji}\rropt{v}{y_{j(k-1)}; w} \\
&+2v_{jk}\ropt{v}{y_{j(k-1)}; w}\Bigr] \\
\rropt{v}{y_k; w} =& \rropt{v}{x_k; w}\sigma_k'(x_k) \\
&+\left(\ropt{v}{x_k; w}\right)^2\sigma_k''(x_k)\end{aligned}$$
The backward computation follows as:
$$\begin{aligned} \rropt{v}{\frac{\partial E}{\partial y_k}; w} =& e_k''(y_k)\left(\ropt{v}{y_k; w}\right)^2 \\
&+e_k'(y_k)\rropt{v}{y_k; w} \\
&+\sum\limits_j \bigg[w_{jk}\rropt{v}{\frac{\partial E}{\partial x_j}; w}+ \\
&2v_{jk}\ropt{v}{\frac{\partial E}{\partial x_j}; w}\bigg] \\
\rropt{v}{\frac{\partial E}{\partial x_k}; w} =&2\ropt{v}{x_k; w}\sigma_k''(x_k)\ropt{v}{\frac{\partial E}{\partial y_k}; w} \\
&+\sigma_k'(x_k)\rropt{v}{\frac{\partial E}{\partial y_k}; w} \\
&+\rropt{v}{x_k; w}\sigma_k''(x_k)\frac{\partial E}{\partial y_k} \\
&+\left(\ropt{v}{x_k; w}\right)^2\sigma_k'''(x_k)\frac{\partial E}{\partial y_k} \\
\rropt{v}{\frac{\partial E}{\partial w_{jk}}; w} =& 2\ropt{v}{y_k; w}\ropt{v}{\frac{\partial E}{\partial x_j}; w} \\
&+y_k\rropt{v}{\frac{\partial E}{\partial x_j}; w} \\
&+\rropt{v}{y_k; w}\frac{\partial E}{\partial x_j}\end{aligned}$$

The original formulation $\ropt{v}{\cdot}$ allows us to efficiently compute $H(w) v$, which can be used to compute $\rho(w)$ and/or estimate the eigenvector $\bar{v}$ corresponding to the spectral radius (via power iteration or LOBPCG). However, the extended formulation $\rropt{v}{\cdot}$ allows us to efficiently compute $v^T \nabla H(w) v$ and thus $\nabla \rho(w)$. This enables us to efficiently compute the gradient of our optimization problem for use in gradient descent methods.

\section{Convergence Analysis Proofs}
\label{A. Analysis}

\subsection{Stochastic Gradient Descent Convergence}
\label{A. SGD Convergence}

First, we prove the convergence of our regularization term (Lemma \ref{epsilon}).

\begin{proof}
We start by splitting $v_k^T\nabla H_k v_k$ into its components
$$\begin{aligned}v_k^T\nabla H_k v_k=&(v_k-\bar{v}_k+\bar{v}_k)^T\nabla H_k (v_k-\bar{v}_k+\bar{v}_k) \\
=&(v_k-\bar{v}_k)^T\nabla H_k (v_k-\bar{v}_k) \\
&+2(v_k-\bar{v}_k)^T\nabla H_k \bar{v}_k+\bar{v}_k^T\nabla H_k \bar{v}_k.\end{aligned}$$
The last term $\bar{v}_k^T\nabla H_k \bar{v}_k=\nabla \bar{\rho}_k$ by definition. We apply the triangle inequality and bind the other terms. Given the convergence criteria on $v_k$ and Assumptions \ref{function} and \ref{lipschitz} (with $||H(w)-H(\omega)||\le L||w-\omega||,\ \forall\ w,\omega\in\R^n, L\ge0$), it follows that
$$\left|\left|(v_k-\bar{v}_k)^T\nabla H_k \bar{v}_k\right|\right|\le L\left|\left|v_k-\bar{v}_k\right|\right| \le L\varepsilon_k.$$
For the first term, we similarly get
$$\begin{aligned}\left|\left|(v_k-\bar{v}_k)^T\nabla H_k (v_k-\bar{v}_k)\right|\right|\le& L\left|\left|v_k-\bar{v}_k\right|\right|^2 \\
\le&L\varepsilon_k^2.\end{aligned}$$
Given Assumption \ref{epsilonK}, the limit
$$\lim\limits_{k\to\infty}v_k^T\nabla H_k v_k=\lim\limits_{k\to\infty}\bar{v}_k^T\nabla H_k \bar{v}_k=\lim\limits_{k\to\infty}\nabla \bar{\rho}_k.$$
\end{proof}

Next, we prove that Algorithms \ref{SGD} and \ref{LOBPCG} follow the assumed update steps (Lemma \ref{Update}):

\begin{proof}
We begin by splitting $p_k$ into its components
$$\begin{aligned}||p_k||^2=&||p_k-\nabla g_k+\nabla g_k||^2 \\
=&||p_k-\nabla g_k||^2+||\nabla g_k||^2 \\
&+2\left(p_k-\nabla g_k\right)^T\nabla g_k.\end{aligned}$$
Let us first assume $\rho_k>K$. By the definition of $g_k$ and the triangle inequality,
$$||\nabla g_k||^2\le||\nabla f_k||^2+2\mu||\nabla f_k||||\nabla \rho_k||+\mu^2||\nabla \rho_k||^2.$$
Taking the expectation (with respect to batch $B_k$ conditioned on the history $\mP_k$) and applying the Cauchy-Schwarz inequality yields
$$\begin{aligned}\E_{B_k}\left[||p_k||^2\big|\mP_k\right]\le &\E_{B_k}\left[||\nabla f_k||^2\big|\mP_k\right] \\
&+2\mu\left(\E_{B_k}\left[||\nabla f_k||^2\big|\mP_k\right]\right)^{\frac{1}{2}} \\
&\times\left(\E_{B_k}\left[||\nabla \rho_k||^2\big|\mP_k\right]\right)^{\frac{1}{2}} \\
&+\mu^2\E_{B_k}\left[||\nabla \rho_k||^2\big|\mP_k\right].\end{aligned}$$
Applying H{\"o}lder's inequality with $||\nabla f_k||^2$, $||\nabla \rho_k||$, $p=3/2$, and $q=3$ yields
$$\begin{aligned}\E_{B_k}\left[||\nabla f_k||^2||\nabla \rho_k||\big|\mP_k\right] \le &\left(\E_{B_k}\left[||\nabla f_k||^3\big|\mP_k\right]\right)^{\frac{2}{3}} \\
&\times\left(\E_{B_k}\left[||\nabla \rho_k||^3\big|\mP_k\right]\right)^{\frac{1}{3}}.\end{aligned}$$
Similarly, using $||\nabla f_k||^3$, $||\nabla \rho_k||$, $p=4/3$, and $q=4$ yields
$$\begin{aligned}\E_{B_k}\left[||\nabla f_k||^3||\nabla \rho_k||\big|\mP_k\right] \le &\left(\E_{B_k}\left[||\nabla f_k||^4\big|\mP_k\right]\right)^{\frac{3}{4}} \\
&\times\left(\E_{B_k}\left[||\nabla \rho_k||^4\big|\mP_k\right]\right)^{\frac{1}{4}}.\end{aligned}$$
Applying this, we obtain
$$\begin{aligned}\E_{B_k}\left[||p_k||^3\big|\mP_k\right]\le&\E_{B_k}\left[||\nabla f_k||^3\big|\mP_k\right] \\
&+3\mu\left(\E_{B_k}\left[||\nabla f_k||^3\big|\mP_k\right]\right)^{\frac{2}{3}} \\
&\times\left(\E_{B_k}\left[||\nabla \rho_k||^3\big|\mP_k\right]\right)^{\frac{1}{3}} \\
&+3\mu^2\left(\E_{B_k}\left[||\nabla f_k||^3\big|\mP_k\right]\right)^{\frac{1}{3}} \\
&\times\left(\E_{B_k}\left[||\nabla \rho_k||^3\big|\mP_k\right]\right)^{\frac{2}{3}} \\
&+\mu^3\E_{B_k}\left[||\nabla \rho_k||^3\big|\mP_k\right],\end{aligned}$$ \\
$$\begin{aligned}\E_{B_k}\left[||p_k||^4\big|\mP_k\right]\le&\E_{B_k}\left[||\nabla f_k||^4\big|\mP_k\right] \\
&+4\mu\left(\E_{B_k}\left[||\nabla f_k||^4\big|\mP_k\right]\right)^{\frac{3}{4}} \\
&\times\left(\E_{B_k}\left[||\nabla \rho_k||^4\big|\mP_k\right]\right)^{\frac{1}{4}} \\
&+6\mu^2\left(\E_{B_k}\left[||\nabla f_k||^4\big|\mP_k\right]\right)^{\frac{1}{2}} \\
&\times\left(\E_{B_k}\left[||\nabla \rho_k||^4\big|\mP_k\right]\right)^{\frac{1}{2}}\\
&+4\mu^3\left(\E_{B_k}\left[||\nabla f_k||^4\big|\mP_k\right]\right)^{\frac{1}{4}} \\
&\times\left(\E_{B_k}\left[||\nabla \rho_k||^4\big|\mP_k\right]\right)^{\frac{3}{4}} \\
&+\mu^4\E_{B_k}\left[||\nabla \rho_k||^4\big|\mP_k\right].\end{aligned}$$

Given Assumption \ref{moments}, this implies that
$$\E_{B_k}\left[||p_k||^j\big|\mP_k\right]\le \overline{A}_j + \overline{B}_j ||w_k||^j,$$
for $j=2,3,4$ and some positive constants $\overline{A}_j, \overline{B}_j$. Combining this with the above results shows that the second, third, and fourth moments of the update term are bounded, as required.

If $p_k\le K$, then $p_k=\nabla f_k$, and the statement follows from Assumption \ref{moments}.
\end{proof}

The rest of this proof uses our assumptions and lemmas and follows \citepos{Bottou98on-linelearning} proof that SGD converges. In the next step, we prove confinement (Lemma \ref{Confinement}).

\begin{proof}
Let $\varphi(x):=\begin{cases} 0, & x<D, \\ (x-D)^2, & x\ge D,\end{cases}$ and $\psi_k := \varphi(||w_k||^2)$. This implies that
$$\varphi(y)-\varphi(x) \le (y-x)\varphi'(x)+(y-x)^2,$$
for $y,x \in \R$. Note that this becomes an equality when $x,y>D$.

Applying this to $\psi_{k+1}-\psi_k$, we derive
$$\begin{aligned}\psi_{k+1}-\psi_k\le &\left(-2\alpha_k w_k^T p_k+\alpha_k^2||p_k||^2\right)\psi'(||w_k||^2)\\
&+4\alpha_k^2\left(w_k^Tp_k\right)^2-4\alpha_k^3w_k^Tp_k||p_k||^2 \\
&+\alpha_k^4||p_k||^4.\end{aligned}$$
By the Cauchy-Schwartz inequality, we get
$$\begin{aligned}\psi_{k+1}-\psi_k\le &-2\alpha_k w_k^T p_k\psi'(||w_k||^2) \\
&+\alpha_k^2||p_k||^2\psi'(||w_k||^2) \\
&+4\alpha_k^2||w_k||^2||p_k||^2+4\alpha_k^3||w_k||||p_k||^3\\
&+\alpha_k^4||p_k||^4.\end{aligned}$$

Taking the expectation, we have
$$\begin{aligned}\E_{B_k}\left[\psi_{k+1}-\psi_k\big|\mP_k\right]\le &-2\alpha_k w_k^T \nabla g_k\psi'(||w_k||^2) \\
&+\alpha_k^2\E_{B_k}\left[||p_k||^2\big|\mP_k\right]\psi'(||w_k||^2)\\
&+4\alpha_k^2||w_k||^2\E_{B_k}\left[||p_k||^2\big|\mP_k\right] \\
&+4\alpha_k^3||w_k||\E_{B_k}\left[||p_k||^3\big|\mP_k\right] \\
&+\alpha_k^4\E_{B_k}\left[||p_k||^4\big|\mP_k\right].\end{aligned}$$
Given Assumption \ref{learning rate}, for sufficiently large $k$, $\alpha_k^2\ge \alpha_k^3 \ge \alpha_k^4$. Due to Lemma \ref{Update}, there exist positive constants $A_0, B_0$ such that
$$\begin{aligned}\E_{B_k}\left[\psi_{k+1}-\psi_k\big|\mP_k\right] \le&-2\alpha_k w_k^T \nabla g_k\psi'(||w_k||^2) \\
&+\alpha_k^2\left(A_0+B_0||w_k||^4\right),\end{aligned}$$
and thus, there exist positive constants $A, B$ such that
$$\begin{aligned}\E_{B_k}\left[\psi_{k+1}-\psi_k\big|\mP_k\right] \le&-2\alpha_k w_k^T \nabla g_k\psi'(||w_k||^2) \\ &+\alpha_k^2\left(A+B\psi_k\right).\end{aligned}$$
If $||w_k||^2<D$, then $\psi'(||w_k||^2)=0$, and the first term on the right-hand side is zero. If $||w_k||^2\ge D$, by Assumption \ref{horizon}, the first term of the right-hand side is negative. Therefore,
$$\E_{B_k}\left[\psi_{k+1}-\psi_k\big|\mP_k\right] \le\alpha_k^2\left(A+B\psi_k\right).$$

We then transform the expectation inequality to
$$\E_{B_k}\left[\psi_{k+1}-(1+\alpha_k^2B)\psi_k\big|\mP_k\right] \le\alpha_k^2A.$$

We define the sequences $\phi_k, \widetilde{\psi}_k$ as follows:
$$\phi_k:=\prod\limits_{i=1}^{k-1}\frac{1}{1+\alpha_i^2B} \text{ and } \widetilde{\psi}_k:=\phi_k\psi_k.$$
Note that $0<\lim\limits_{k\to\infty} \phi_k:=\phi_\infty<\infty$ (this can be shown by considering $\log\phi_k$ and using the condition on the sum of the squared learning rate). By substituting these sequences into the above inequality, we obtain
$$\E_{B_k}\left[\widetilde{\psi}_{k+1}-\widetilde{\psi}_k\big|\mP_k\right] \le\alpha_k^2\phi_{k+1}A.$$

By defining $\delta_k(u) := \left(\E\left[u_{k+1}-u_k\right]\right)^+$, for some process $u_k$, we can bound the positive expected variations of $\widetilde{\psi}_k$, as follows
$$\begin{aligned}\E\left[\delta_k\left(\widetilde{\psi}\right)\right]=&\E\left[\left(\E_{B_k}\left[\widetilde{\psi}_{k+1}-\widetilde{\psi}_k\big|\mP_k\right]\right)^+\right] \\
\le&\alpha_k^2\phi_{k+1}A.\end{aligned}$$
Due to Assumption \ref{learning rate}, the sum of this expectation is finite. By the Quasi-Martingale Convergence Theorem, $\widetilde{\psi}_k$ converges almost surely. And, since $\phi_k$ converges to $\phi_\infty>0$, $\psi_k$ converges almost surely.
Suppose $\lim\limits_{k\to\infty} \psi_k = \psi_\infty > 0$.

If $\{w_k\}_{k=1}^\infty$ is unbounded, then for sufficiently large $k \ge \kappa$, $||w_k||^2 > D+1$ and $\psi'(||w_k||^2) \ge c_1 > 0$. Without loss of generality, we assume this instead of dealing with a subsequence. Under these conditions, the given inequality becomes equality
$$\begin{aligned}\psi_{k+1}-\psi_k=&\left(-2\alpha_k w_k^T p_k+\alpha_k^2||p_k||^2\right)\psi'(||w_k||^2)\\
&+\left(-2\alpha_k w_k^T p_k+\alpha_k^2||p_k||^2\right)^2.\end{aligned}$$
Therefore, we can express $\psi_\infty$ as the infinite sum
$$\begin{aligned}\psi_\infty- \psi_\kappa = &\sum\limits_{k=\kappa}^\infty\left[\psi_{k+1}-\psi_k\right] \\
=&\sum\limits_{k=\kappa}^\infty\Big[\left(-2\alpha_k w_k^T p_k+\alpha_k^2||p_k||^2\right)\psi'(||w_k||^2)\\
&+\left(-2\alpha_k w_k^T p_k+\alpha_k^2||p_k||^2\right)^2\Big].\end{aligned}$$

The next statements hold almost surely. We have 
$$\sum\limits_{k=\kappa}^\infty \left(-2\alpha_k w_k^T p_k+\alpha_k^2\|p_k\|^2\right)^2\le\sum\limits_{k=\kappa}^\infty \alpha_k^2\left(A+B\psi_k\right).$$
This can be seen by expanding the square and using Cauchy-Schwarz and Lemma \ref{Update}.
Since $\psi_k$ converges almost surely, it is bounded above by $\psi_k\le c_2$ almost surely. Defining $c_3 := A+Bc_2$, we have
$$\sum\limits_{k=\kappa}^\infty\left(-2\alpha_k w_k^T p_k+\alpha_k^2||p_k||^2\right)^2 \le\sum\limits_{k=\kappa}^\infty \alpha_k^2c_3.$$\\
Assumption \ref{learning rate} implies the convergence of the series on the right-hand side. Because the terms of the sum on the left are non-negative, this sum also converges almost surely by the Monotone Convergence Theorem. Similarly, the sum $\sum_{k=\kappa}^\infty\alpha_k^2||p_k||^2\psi'(||w_k||^2)$ converges almost surely. This uses Assumption \ref{learning rate}, Lemma \ref{Update}, and that $\psi_k$ and $\psi'(||w_k||^2)$ are positive and almost surely bounded above.

Now, we subtract these convergent series from the equation for $\psi_\infty$ to get
$$\begin{aligned}
&\psi_\infty- \psi_\kappa - \sum\limits_{k=\kappa}^\infty\alpha_k^2\|p_k\|^2\psi'(\|w_k\|^2) \\
&-\sum\limits_{k=\kappa}^\infty\left(-2\alpha_k w_k^T p_k+\alpha_k^2\|p_k\|^2\right)^2\\
=&\sum\limits_{k=\kappa}^\infty\Big[\left(-2\alpha_k w_k^T p_k+\alpha_k^2\|p_k\|^2\right)\psi'(\|w_k\|^2) \\
&+ \left(-2\alpha_k w_k^T p_k+\alpha_k^2\|p_k\|^2\right)^2\Big] \\
&-\sum\limits_{k=\kappa}^\infty\alpha_k^2\|p_k\|^2\psi'(\|w_k\|^2) \\
&-\sum\limits_{k=\kappa}^\infty\left(-2\alpha_k w_k^T p_k+\alpha_k^2\|p_k\|^2\right)^2.
\end{aligned}$$
Since the involved series almost surely converge, we can combine the terms to obtain
$$\begin{aligned}
&\psi_\infty- \psi_\kappa - \sum\limits_{k=\kappa}^\infty \Big[ \alpha_k^2\|p_k\|^2\psi'(\|w_k\|^2) \\
&+ \left(-2\alpha_k w_k^T p_k+\alpha_k^2\|p_k\|^2\right)^2\Big] \\
=&\sum\limits_{k=\kappa}^\infty-2\alpha_k w_k^T p_k\psi'(\|w_k\|^2).\end{aligned}$$

Let us consider the sum on the right-hand side. By Assumption \ref{horizon}, $w_k^T \nabla g_k \ge c_4 > 0$ and thus
$$\sum\limits_{k=\kappa}^\infty-2\alpha_k w_k^T p_k\psi'(\|w_k\|^2)\le\sum\limits_{k=\kappa}^\infty-2\alpha_k c_4 c_1.$$
Assumption \ref{learning rate} implies that this sequence diverges to negative infinity. This yields a contradiction, since the left-hand side must be a finite value. Therefore, $\{w_k\}_{k=1}^\infty$ must be bounded.
\end{proof}

Next, we prove that SGD converges almost surely (Theorem \ref{Convergence}).

\begin{proof}
All statements here are taken almost surely. By Assumption \ref{function}, we have $f\in\C^5$. From linear algebra, the Hessian of $\rho(w)$ is continuous (the largest eigenvalue is a continuous function of a parametric matrix with continuous functions). This implies that $g\in C^3$, and thus, by Lemma \ref{Confinement} it is bounded on the set of all iterates. We can bound differences in the loss criteria $g_k$ using a first-order Taylor expansion and bounding the second derivatives with $K_1$.
$$|g_{k+1}-g_k+\alpha_kp_k^T\nabla g_k|\le \alpha_k^2 ||p_k||^2 K_1.$$
This can be rewritten as:
$$g_{k+1}-g_k\le-\alpha_kp_k^T\nabla g_k+ \alpha_k^2 ||p_k||^2 K_1.$$
Taking the expectation, we get
$$\begin{aligned}\E_{B_k}\left[g_{k+1}-g_k\big|\mP_k\right]\le&-\alpha_k\E_{B_k}\left[p_k^T\nabla g_k\big|\mP_k\right] \\
&+ \alpha_k^2 \E_{B_k}\left[||p_k||^2\big|\mP_k\right] K_1.\end{aligned}$$
We decompose $p_k=\nabla g_k + (p_k-\nabla g_k)$ and bound the expectation using Lemmas \ref{Update} and \ref{Confinement}
$$\E_{B_k}\left[||p_k||^2\big|\mP_k\right]\le A_2 + B_2 ||w_k||^2\le K_2.$$
This yields
\begin{equation}\label{eq:g}
    \begin{aligned}\E_{B_k}\left[g_{k+1}-g_k\big|\mP_k\right]\le-\alpha_k||\nabla g_k||^2 +\alpha_k^2 K_1K_2\\
    -\alpha_k\E_{B_k}\left[(p_k-\nabla g_k)^T\nabla g_k\big|\mP_k\right].\end{aligned}
\end{equation}
Next, we apply the Cauchy-Schwarz inequality and bound the error term. From our proof to Lemma \ref{epsilon}, we have
$$\left|\left|v_k^T\nabla H_k v_k-\bar{v}_k^T\nabla H_k \bar{v}_k\right|\right|\le 2L\varepsilon_k = C_1\varepsilon_k.$$
This implies 
$$\left|\left|\E_{B_k}\left[p_k\big|\mP_k\right]-\nabla g_k\right|\right|\le C_1\varepsilon_k.$$
We also bound $||\nabla g_k||\le C_2$ using Lemma \ref{Confinement}.
Combining these bounds yields
\begin{equation}\label{eq:error}
    \left|\left|\E_{B_k}\left[p_k\big|\mP_k\right]-\nabla g_k\right|\right| ||\nabla g_k||\le \varepsilon_k K_3.
\end{equation}
Applying \eqref{eq:error} to \eqref{eq:g} gives us
$$\E_{B_k}\left[g_{k+1}-g_k\big|\mP_k\right]\le \alpha_k^2 K_1 K_2 + \alpha_k\varepsilon_k K_3.$$
The positive expected differences are then bounded by
$$\begin{aligned}\E_{B_k}\left[\delta_k\left(h\right)\big|\mP_k\right]=&\E_{B_k}\left[\delta\E_{B_k}\left[g_{k+1}-g_k\big|\mP_k\right]\right]\\
\le&\alpha_k^2 K_1 K_2 + \alpha_k\varepsilon_k K_3.\end{aligned}$$
By the Quasi-Martingale Convergence Theorem, $g_k$ converges almost surely,
$$g_k \xrightarrow[k\to\infty]{\text{a.s.}} g_\infty.$$

Since $g_k$ converges, $\sum_{k=1}^\infty\E_{B_k}\left[g_{k+1}-g_k\big|\mP_k\right]$ also converges. Furthermore, the series $\sum_{k=1}^\infty \alpha_k^2K_1K_2$ and $\sum_{k=1}^\infty\alpha_k\varepsilon_k K_3$ converge due to Assumption \ref{learning rate}. From \eqref{eq:g} we have
\begin{equation}\label{eq:nabg}
    \sum\limits_{k=1}^{\infty}\alpha_k||\nabla g_k||^2<\infty.
\end{equation}
We define $\theta_k=||\nabla g_k||^2$. The differences of $\theta_k$ are bounded using the Taylor expansion, similarly to the differences of $g_k$
$$\theta_{k+1}-\theta_k\le -2\alpha_kp_k^T\nabla^2 g_k\nabla g_k+\alpha_k^2||p_k||^2 K_4,$$
for some constant $K_4$. Taking the expectation, we decompose $p_k$ and bound $||p_k||^2$ similarly to \eqref{eq:g}. 
$$\begin{aligned}\theta_{k+1}-\theta_k\le -2\alpha_k\nabla g_k^T\nabla^2 g_k\nabla g_k+\alpha_k^2K_2K_4 \\
-2\alpha_k\E_{B_k}\left[(p_k-\nabla g_k)^T\nabla^2 g_k\nabla g_k\big|\mP_k\right]\end{aligned}$$
We also bound the second derivative by $||\nabla^2 g_k||\le K_5/2$ and the error term using \eqref{eq:error}, yielding
$$\begin{aligned}\E_{B_k}\left[\theta_{k+1}-\theta_k\big|\mP_k\right]\le &\alpha_k||\nabla g_k||^2K_5+\alpha_k^2K_2K_4 \\
&+\alpha_k\varepsilon_kK_3K_5.\end{aligned}$$
The positive expectations are bounded,
$$\begin{aligned}\E_{B_k}\left[\delta_k\left(\theta\right)\big|\mP_k\right]=&\E_{B_k}\left[\delta\E_{B_k}\left[\theta_{k+1}-\theta_k\big|\mP_k\right]\right] \\
\le&\alpha_k||\nabla g_k||^2K_5+\alpha_k^2K_2K_4 \\
&+\alpha_k\varepsilon_kK_3K_5.\end{aligned}$$
Since the terms on the right-hand side are sums of convergent infinite sequences (due to Assumption \ref{learning rate} and \eqref{eq:nabg}), by the Quasi-Martingale Convergence Theorem, $\theta_k$ converges almost surely. Suppose $||\nabla g_k||$ converges to a positive value $C_3>0$. Then for sufficiently large $k\ge\kappa$ there exists a positive constant $0<C_4<C_3$ such that $||\nabla g_k||\ge C_4$. Thus, $\sum\limits_{k=\kappa}^{\infty}\alpha_k||\nabla g_k||^2 \ge C_4^2 \sum\limits_{k=\kappa}^{\infty}\alpha_k$. By Assumption \ref{learning rate}, this diverges, contradicting \eqref{eq:nabg}. Therefore, the limit must be zero
$$\theta_k \xrightarrow[k\to\infty]{\text{a.s.}} 0 \text{ and } \nabla g_k \xrightarrow[k\to\infty]{\text{a.s.}} 0.$$
\end{proof}

\section{Additional Experiment Details}
\label{A. Experiments}

The code is available at \href{https://anonymous.4open.science/r/spectral-radius/}{https://anonymous.4open.science/r/spectral-radius/}. The algorithm is written in Python, using PyTorch \citep{pytorch} and TorchVision \citep{torchvision}. Forest cover-type and USPS experiments are run on a 3.1 GHz Dual-Core Intel Core i5 processor with 16 GB 2133 MHz LPDDR3 memory. Chest X-ray experiments are run on an Intel Xeon CPU E5-2650 v4 @ 2.20GHz with an NVIDIA Tesla K40c GPU.

\subsection{Data Sets}
\label{A. Data}

The forest cover-type data \citep{cov} uses cartographic data to predict the tree species (as determined by the United States Forest Service) of a 30 x 30-meter cell. This cartographic data includes elevation, aspect, slope, distance to surface water features, distance to roadways, hill-shade index at three times of day, distance to wildfire ignition points, wilderness area designation, and soil type. Seven major tree species are included: spruce/fir, lodgepole pine, Ponderosa pine, cottonwood/willow, aspen, Douglas-fir, and krummholz. In total, 581,012 samples are included, which we split 64\%/16\%/20\% into train/validation/test data sets.

The USPS digits data \citep{USPS} includes 16 x 16 pixel greyscale images from scanned envelopes to identify which digit 0-9 each image corresponds to. This data set is already split into 7,291 training and 2,007 test images. We take 1/7 of the training set as validation.

The chest X-ray data \citep{chestxray} contains 1024 x 1024 pixel color images of patients' chest X-rays, to identify which of fourteen lung diseases each patient has. Note that this is a multi-label problem; patients can have none, one, or multiple of these conditions. A total of 112,120 patients' images are included, taken between the years 1992 and 2015, which we split 70\%/10\%/20\% into train/validation/test data sets.

\subsection{Generalization Tests}
\label{A. Generalization}

For forest cover-type data, we weight the test subjects to shift the mean of a feature or multiple features. Since we normalize the data, the weight of each test subject is determined by the ratio of the normal probability distribution function value with and without the shift. This shift adds a slight bias to the test set that is not in the original data set. We first use this shift method to increase the mean of each feature value by 0.1, compare the accuracy of our trained models, and find that certain features are problematic. Upon further examination, this is because these features are binary factors with rare classes (so our weighting of subjects emphasizes a few of them). Then, we shift each feature (except the problematic features) by a random normal amount (with mean 0 and standard deviation 0.05), compare the accuracy of our trained models, and repeat it one thousand total times.

For USPS handwritten digits data, we augment the test set: Augmented Test 1 uses random crops (with padding) of up to one pixel and random rotations of up to 15$^\circ$; Augmented Test 2 uses crops of up to two pixels and rotations of up to 30$^\circ$. Note that we do not augment our training set while learning our models, as a similar augmentation would yield comparable training and test sets.

For the Conditional GAN examples, we modify \citepos{Linder-Noren} implementation and generate 10,000 images from the trained generator model. We train the GAN model with a batch size of 64, cosine annealing learning rate (initially $10^{-4}$), $\beta_1=0.5$, and $\beta_2=0.999$. We randomly smooth the labels to be uniform between 0.0 and 0.3 for generated samples and 0.7 and 1.0 for true samples. We also swap the labels on 1\% of batches, chosen at random. We also modify \citepos{Chhabra} implementation and generate 10,000 images from the trained generator model.

For the chest X-ray data, we compare performance on two similar data sets, CheXpert \citep{CheXpert} and MIMIC-CXR \citep{MIMIC-CXR}. For this comparison, we only consider the six conditions common to the three data sets: atelectasis, cardiomegaly, consolidation, edema, pneumonia, and pneumothorax. Additionally, we ignore any uncertain labels in the CheXpert and MIMIC-CXR classes. We keep the assigned training and validation data sets separate for each data set, as there appear to be differences in labeling. Particularly, the CheXpert validation set is fully labeled, while the training set contains uncertain and missing labels. While these are labeled "training" and "validation" sets, we solely use them as test sets. CheXpert contains 234 validation and 223,415 training images from Stanford Hospital, taken between 2002 and 2017. MIMIC-CXR contains 2,732 validation and 369,188 training images from Beth Israel Deaconess Medical Center, taken between the years 2011 and 2016.

We measure the spectral radius $\rho$ of each model on the full training set. We use $\varepsilon=10^{-3}$ and a maximum of 1,000 power iterations, except for the chest X-ray models, where we use $\varepsilon=0.1$ and a maximum of 100 power iterations. These values allow the algorithm to find an accurate eigenvalue within a reasonable run time.

\subsection{Implementation}
\label{A. Implementation}

For forest cover-type models, we train using stochastic gradient descent, with learning rate $\frac{0.5}{\text{epoch \#}}$, batch size of 128, and a maximum of 100 epochs. For models with batch sizes 32 and 64, we use a $\frac{0.1}{\text{epoch \#}}$ learning rate instead. The network uses 3 hidden layers with 20 hidden nodes in each layer. The learning rate and maximum number of epochs allow our algorithm to converge to an accurate model. We experiment with other feed-forward networks, learning rates, and optimizers but find that this structure works best. Our experiments with batch size are discussed in Section \ref{batch size}.

For USPS models, we use the Adam optimizer with a learning rate of $10^{-3}$, a batch size of 128, and a maximum of 100 epochs. The network takes an input image and processes it through the following layers in order: convolution to 8 channels, pool, convolution to 16 channels, pool, convolution to 32 channels, pool, fully connected to 64 nodes, fully connected to 10 nodes, softmax. All convolutions are of kernel size 3, stride 1, and padding 1; all pools are max pooling with kernel size 2 and stride 2. ReLUs are used to connect the various layers. The MNIST images are re-sized to be 16 x 16 to match the model's input size. We similarly experiment with feed-forward and other convolutional networks, learning rates, and optimizers, but find that this structure performs best.

For the chest X-ray models, we resize the images to 256 x 256 and then crop them to 224 x 224. We use the Adam optimizer with a learning rate of $10^{-5}$, batch size of 4, a maximum of 100 power iterations, random initialization of each power iteration, and gradient clipping (at a magnitude of 100). The crops and small batch size are necessary for the GPUs on our server to not run out of memory (12 GB). Using a CheXNet model \citep{CheXNet} (trained using their methodology) as the initialization, we train for an additional epoch, except for the Asymmetric Valley model (which we address later). We also try 5 epochs; however, the first epoch has the highest validation mean AUC in each case.

The entropy-SGD models are trained with a learning rate of 0.1 (except on chest X-ray data, where 0.001 is used), a momentum of 0.9, and no dampening or weight decay. The K-FAC models are trained using \citepos{alecwangcq} implementation, with a learning rate of 0.001 (except on chest X-ray data, where $10^{-7}$ is used) and a momentum of 0.9.

The Asymmetric Valley models are trained with an initial learning rate of 0.5 for 250 epochs (iterations 161-200 utilizing SWA). For chest X-ray data, we start at the SWA step, using CheXNet initialization \citep{CheXNet}.

The forest cover-type LOBPCG model is trained with regularization parameters $\mu=.0028$ and $K=1$, update frequency $b=4$ and learning rate $\tilde{\alpha}(j) = \exp(-4j-2)$. The USPS LOBPCG model is trained with regularization parameters $\mu=.005$ and $K=0$, update frequency $b=4$ and learning rate $\tilde{\alpha}(j) = \exp(-4j)$.

\section{Constructed Data Sets}
\label{Constructed}

\begin{figure*}[t]
    \centering
    \begin{subfigure}[t]{.49\textwidth}
    \centering
    \includegraphics[scale=0.49]{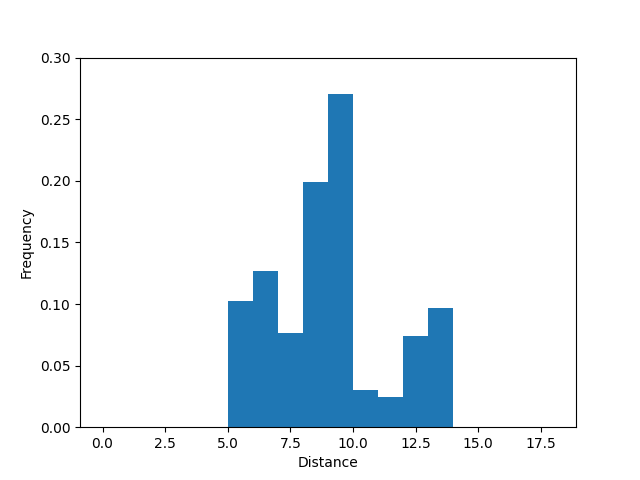}
    \caption{GAN1}
    \label{GAN-Dist}
    \end{subfigure}
    \begin{subfigure}[t]{.49\textwidth}
    \centering
    \includegraphics[scale=0.49]{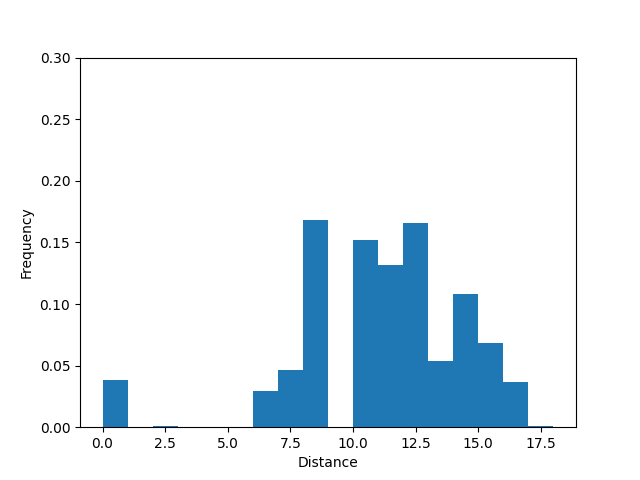}
    \caption{Const1}
    \label{Const1-Dist}
    \end{subfigure}
    \begin{subfigure}[t]{.49\textwidth}
    \centering
    \includegraphics[scale=0.49]{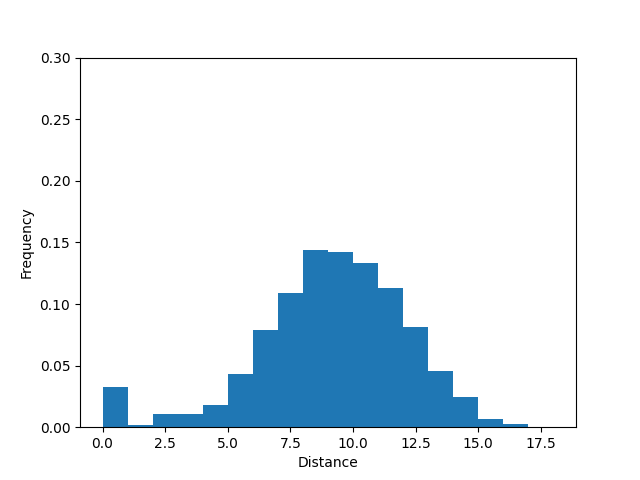}
    \caption{AT1}
    \label{AT1-Dist}
    \end{subfigure}
    \begin{subfigure}[t]{.49\textwidth}
    \centering
    \includegraphics[scale=0.49]{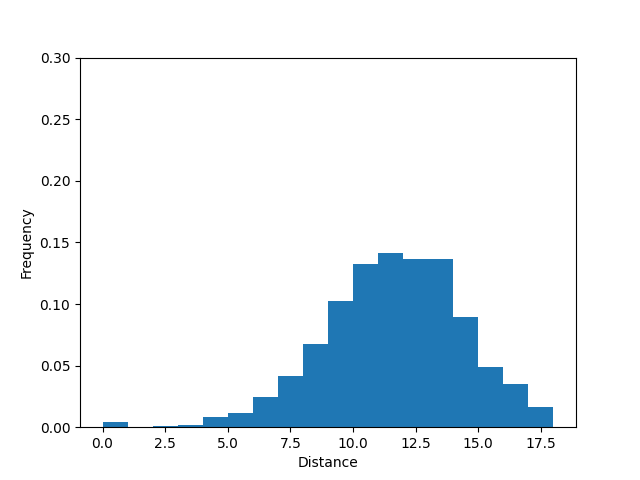}
    \caption{AT2}
    \label{AT2-Dist}
    \end{subfigure}
    \caption{Histograms of Euclidean distance between data sets and USPS test data}
    \label{Histograms-Dist}
\end{figure*}

\begin{figure*}[t]
    \begin{subfigure}[t]{.49\textwidth}
    \centering
    \includegraphics[scale=0.49]{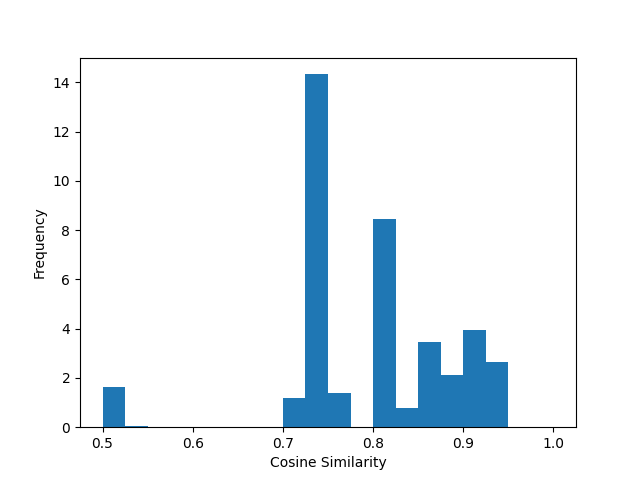}
    \caption{GAN1}
    \label{GAN-Cos}
    \end{subfigure}
    \begin{subfigure}[t]{.49\textwidth}
    \centering
    \includegraphics[scale=0.49]{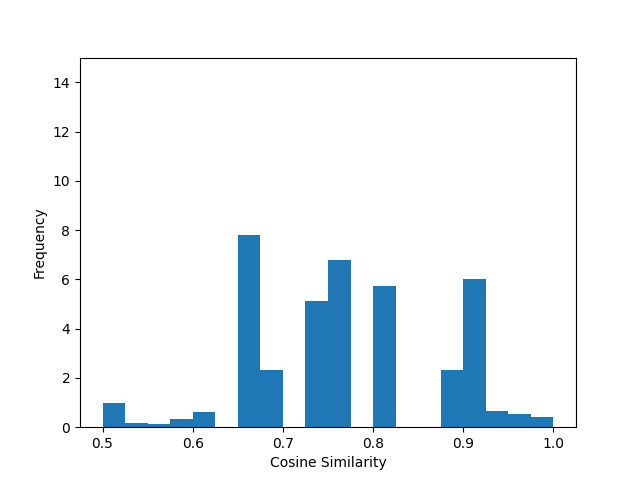}
    \caption{Const2}
    \label{Const2-Cos}
    \end{subfigure}
    \begin{subfigure}[t]{.49\textwidth}
    \centering
    \includegraphics[scale=0.49]{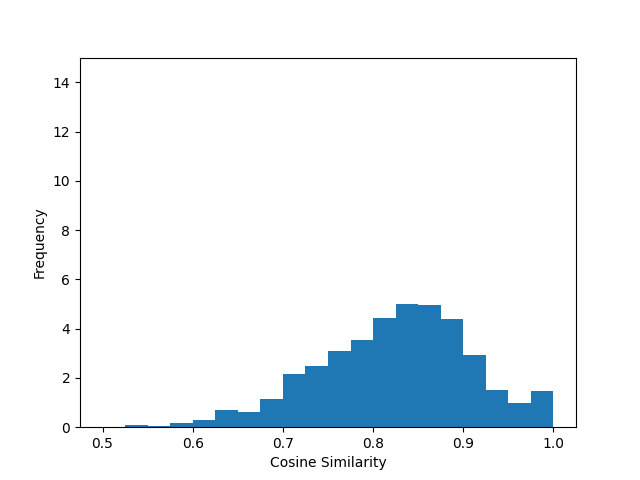}
    \caption{AT1}
    \label{AT1-Cos}
    \end{subfigure}
    \begin{subfigure}[t]{.49\textwidth}
    \centering
    \includegraphics[scale=0.49]{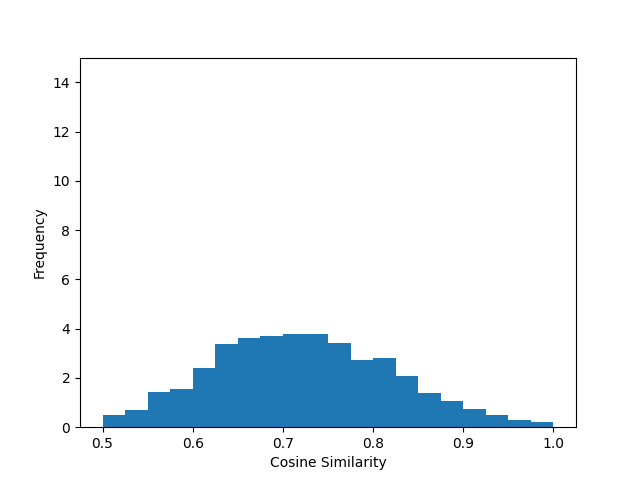}
    \caption{AT2}
    \label{AT2-Cos}
    \end{subfigure}
    \caption{Histograms of cosine similarity between data sets and USPS test data}
    \label{Histograms-Cos}
\end{figure*}

During analysis of the performance of GAN1, we compute the minimum distance ($L_2$-norm) between each image in the GAN1 data set and the images in the USPS test data. We notice that the GAN1 images are distributed differently, relative to the USPS test data, compared to the other data sets. In particular, Figure \ref{Histograms-Dist} shows that the augmented data sets have a bell-curve distribution of such distances, while GAN1 has an abnormal distribution. We construct a data set, Const1, from the augmented test data sets using the following procedure.

\begin{enumerate}
    \item Split the data with respect to distances into integer bins [0,1), [1,2), $\cdots$, [17,18).
    \item Uniformly at random, select 5 bins to draw zero images from.
    \item For the remaining bins, select one of the two augmented test data sets at uniform random. Add all images from the selected data set in the bin's distance range to the constructed data set.
\end{enumerate}

This procedure creates a constructed data set intended to emulate the abnormal distribution of the GAN1 data.

We repeat this process with maximum cosine similarity between images and observe similar distributional abnormalities in the GAN1 data (Figure \ref{Histograms-Cos}). We construct Const2 from the augmented data set using a similar procedure, but with bins [0.5, 0.525), [0.525, 0.55), $\cdots$, [0.975, 1.0).

\section{Grad-CAM}
\label{Grad-CAM}

\begin{table*}[t]
\begin{tabular}{l|r|r|r|r|r|r|r|r}
 & & \multicolumn{6}{c|}{Jaccard Score of Overlap on CheXpert Validation Data} & \\ \cline{3-8}
Model & $\rho$ & SpecRad & EntropySGD & UnReg & KFAC & AsymValley & Mean & PerfDrop \\ \hline \hline
SpecRad & 38.92 & 1.000 & 0.508 & 0.192 & 0.294 & 0.374 & 0.342 & -4.15\% \\
EntropySGD & 41.11 & 0.508 & 1.000 & 0.181 & 0.280 & 0.435 & 0.351 & -3.09\% \\
UnReg & 68.29 & 0.192 & 0.181 & 1.000 & 0.283 & 0.152 & 0.202 & -10.30\% \\
KFAC & 84.65 & 0.294 & 0.280 & 0.283 & 1.000 & 0.261 & 0.280 & -13.16\% \\
AsymValley & 1198.69 & 0.374 & 0.435 & 0.152 & 0.261 & 1.000 & 0.306 & -10.06\%
\end{tabular}
\caption{There is more overlap in the explanations from the two models with low spectral radius $\rho$ (our spectral radius regularization and entropy-SGD) on the CheXpert Validation data set.}
\label{CheXpert Validation Jaccard}
\end{table*}

\begin{table*}[t]
\begin{tabular}{l|r|r|r|r|r|r|r|r}
 & & \multicolumn{6}{c|}{Jaccard Score of Overlap on MIMIC-CXR Validation Data} & \\ \cline{3-8}
Model & $\rho$ & SpecRad & EntropySGD & UnReg & KFAC & AsymValley & Mean & PerfDrop \\ \hline \hline
SpecRad & 38.92 & 1.000 & 0.504 & 0.216 & 0.285 & 0.313 & 0.330 & -8.10\% \\
EntropySGD & 41.11 & 0.504 & 1.000 & 0.172 & 0.229 & 0.358 & 0.316 & -9.21\% \\
UnReg & 68.29 & 0.216 & 0.172 & 1.000 & 0.272 & 0.134 & 0.199 & -10.80\% \\
KFAC & 84.65 & 0.285 & 0.229 & 0.272 & 1.000 & 0.192 & 0.245 & -10.71\% \\
AsymValley & 1198.69 & 0.313 & 0.358 & 0.134 & 0.192 & 1.000 & 0.249 & -10.84\%
\end{tabular}
\caption{There is more overlap in the explanations from the two models with low spectral radius $\rho$ on the MIMIC-CXR Validation data set.}
\label{MIMIC-CXR Validation Jaccard}
\end{table*}

We use \citepos{jacobgilpytorchcam} Grad-CAM implementation to highlight which areas of the chest X-rays are important to our best regularized model ($\mu=10^{-4}$ and $\alpha=10^{-6}$) and the predictions of the baseline models. We compute the Jaccard index of the top 10\% of pixels in each Grad-CAM image to compare which regions are important to each model.

Tables \ref{CheXpert Validation Jaccard} and \ref{MIMIC-CXR Validation Jaccard} show that the two models with the lowest spectral radius $\rho$, our regularized model and entropy-SGD, have the highest overlap in explanations. These models highlight similar areas of the chest X-rays as important in making predictions. The Jaccard scores of their overlap are over .5 on the CheXpert and MIMIC-CXR validation data, the highest of any pair of models. Since these models also perform best on these transfer learning data, evidence suggests that models with low spectral radius generalize better in both their explanations and predictions. The three models with higher spectral radii have a larger dip in performance and less overlap in their explanations.

In contrast, the three models with higher spectral radius, unregularized, K-FAC, and asymmetric valley, have a higher performance drop on the transfer learning data sets and have less overlap in their explanations. These models have mean Jaccard scores of .202-.306 on the CheXpert Validation data, lower than the spectral radius regularization and EntropySGD scores of .342 and .351. The high spectral radius models have scores of .199-.249 on MIMIC-CXR Validation data, while the low spectral radius models have scores of .330 and .316.

\end{document}